\documentclass[review]{elsarticle}

\usepackage{lineno,hyperref}
\usepackage{graphics}
\graphicspath{}
\usepackage{amsmath}
\usepackage{verbatim}
\usepackage{enumerate}
\usepackage{color}

\usepackage{algorithm}  
\usepackage{algpseudocode}  
\usepackage{amsmath}  
\usepackage{caption} 
\usepackage{array}
\usepackage{txfonts} 
\usepackage{ragged2e}
\usepackage{url}
\usepackage{amssymb}
\usepackage{array} 

\usepackage{booktabs} 

\usepackage{longtable}

\usepackage{bm}

\usepackage{setspace} 

\usepackage{natbib} 

\begin{document}
\captionsetup[figure]{labelfont={bf},name={Fig.},labelsep=period}
\captionsetup[table]{labelfont={bf},name={Table.},labelsep=period}
\begin{frontmatter}

\title{Non-linear and Selective Fusion of Cross-Modal Images}
\author{Aiqing Fang, Xinbo Zhao*, Jiaqi Yang, Yanning Zhang, Xiang Zheng}%
\address{
	\justifying\let\raggedright\justifying
National Engineering Laboratory for Integrated Aero-Space-Ground-Ocean Big Data Application Technology, School of Computer Science and Engineering, Northwestern Polytechnical University, Xi’an 710072, China

}



\cortext[mycorrespondingauthor]{Corresponding author}
\ead{xbozhao@nwpu.edu.cn (X. Zhao); aiqingf@mail.nwpu.edu.cn (A. Fang);
	jqyang@nwpu.edu.cn;
	 ynzhang@nwpu.edu.cn}

\begin{abstract}
The human visual perception system has strong robustness in many computer vision tasks. This robustness is closely related to the feature selection and non-linear characteristics of human visual perception system. In order to simulate the human visual perception mechanism in image fusion tasks, we \textit{propose a cross-modal image fusion method that combines illuminance factors and attention mechanisms.} The framework effectively combines traditional image features and modern deep learning features. \textbf{Firstly}, in order to avoid high and low frequency mixing and reduce halo effect, we perform cross-modal image multi-scale decomposition. \textbf{Secondly}, in order to remove highlights, the visual saliency map and the deep feature map are combined with the illuminance fusion factor to perform high-low frequency non-linear fusion. \textbf{Thirdly}, the characteristics of high and low frequency fusion are selected through the channel attention network to obtain the final fusion map. By simulating the non-linear characteristics and selection characteristics of the human visual perception system in image fusion, the fused image is more in line with the human visual perception mechanism. \textbf{Finally}, we validate our fusion framework on public datasets of infrared and visible images and medical images. \textit{The experimental results demonstrate the superiority of our fusion method in visual quality and robustness.} 
\end{abstract}

\end{frontmatter}

\section{Introduction}

Robustness of image fusion has always been a bottleneck problem that puzzles and restricts the application and popularization of traditional image fusion technology, while human beings have strong robustness in many computer vision tasks, such as object detection, object recognition and image caption, et al. \textit{So, we believe that human beings should also have strong robustness in the field of cross-modal image fusion task.} From the perspective of cognitive psychology, the human visual perception system has \textit{the characteristics of information selection} for the perception of external stimuli, and {\textit{the human brain has non-linear characteristics}} for the fusion of perceptual information \cite{Treisman1980A,Valois1996Visual,Kubovy2003Foundations,Gallistel2009Memory}. The \textit{visual attention model} based on feature selection and \textit{convolution neural network} based on non-linear characteristics of brain neuroscience have achieved remarkable results in many computer vision fields. So, we also believe that these two characteristics of human visual perception have positive significance to improve the robustness of cross-modal image fusion task, as verified in Sect.4. 

\begin{figure}[ht]
	
	
	\centering
	\includegraphics[width=1\textwidth]{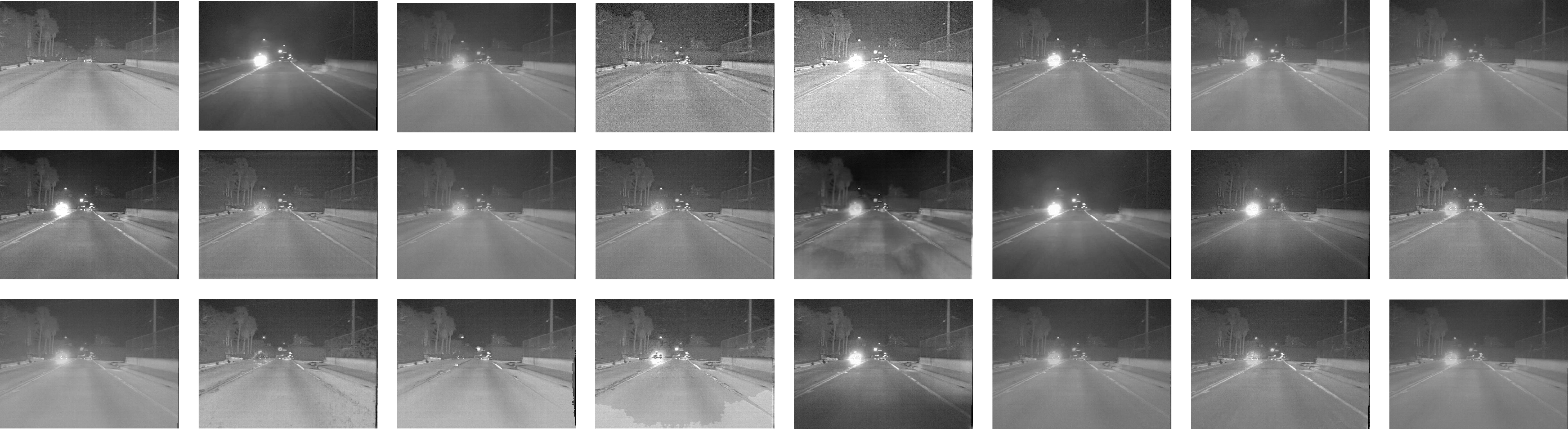}
	\caption{  Schematic illustration of image fusion. From left to right: IR image\cite{FLIR}, Visible image\cite{FLIR}, JSR\cite{Zhang2013Dictionary},  OURS, WLS\cite{Ma2017InfraredWLS},    DDLatLRR\cite{Liu2011LatentLATLRR}, LATLRR\cite{Li2018InfraredLTLRR}, ZCA\cite{Li2018Infrared}, CNN\cite{Liu2017InfraredCNN}, CVT\cite{Nencini2007RemoteCVT}, DL\cite{Li_2018DL}, DTCWT\cite{Liu2015MultiDSIFT}, FusionGAN\cite{MaFusionGAN}, GF\cite{Shutao2013ImageGF}, GTF\cite{Ma2016InfraredGTF}, LP\cite{Burt1987TheLP}, FEZ\cite{Lahoud2019FastZERO}, CBF\cite{Shreyamsha2015ImageCBF}, CSR\cite{Liu2016ImageCSR}, JSRD\cite{Liu2017InfraredJSR-SD}, LP-SR\cite{Liu2015ALPSR}, MSVD\cite{Naidu2011Image}, RP\cite{Toet1989ImageRP}, Wavelet\cite{Chipman1995Wavelets}. 
		Our method has a good fusion effect for high light, and the fusion effect is more coincident with human visual perception mechanism. 
	}
	
	\label{summary}
\end{figure}

In the past few decades, researchers have proposed many image fusion methods based on human visual perception characteristics. For example, a multi-scale decomposition method based on the sensitivity of human eyes to different brightness regions \cite{Li2011Performance}, a convolutional neural network method inspired by neurobiology \cite{Xiang2015A} and a saliency method based on human visual attention mechanism \cite{ZhangInfrared}. Among them, multi-scale decomposition focuses more on hierarchical feature extraction of images. The method based on convolution neural network focuses more on learning the characteristics of images by data driven. The method based on visual saliency focuses more on feature extraction of saliency feature map or saliency object. \textbf{1) In fusion criteria}, the above methods generally use weighted average, maximum or principal component method \cite{Ma2018Infrared} in image fusion criteria, and those on \textit{non-linear feature fusion are few}. \textbf{2) In feature selection}, more emphasis is placed on the extraction and selection of origin image features in the early stage, and the effective selection of \textit{fusion features is lacking}. \textit{However, the human visual fusion perception system is a highly complex non-linear system. Complex characteristics are not only reflected in feature extraction, but also in image information fusion of human brain} \cite{Treisman1980A,Kubovy2003Foundations,Gallistel2009Memory}. In the task of cross-modal image fusion, the human brain filters the perceived target features based on subjective intention, ignores uncertain signals, and fuses non-mutually exclusive features according to prior knowledge. \textit{In order to make the results of cross-modal image fusion more in line with the human visual perception system, narrow the gap between human visual perception system and cross-modal image fusion, we propose a non-linear and selective fusion of cross-modal image based on cognitive psychology theory}. We give a representative example in Figure \ref{summary} to demonstrate the superiority of our method over existing mainstream algorithms. The image is a thermal infrared and visible image, and the data is derived from the traffic dataset in FLIR \cite{FLIR}. From the fused image, we can clearly find that there is a strong boundary effect in the glare. At present, mainstream image fusion algorithms cannot effectively remove glare. Our framework can effectively remove glare regions by introducing non-linear illuminance influence factors, and the fused images have higher clarity.

Our method is not a simple superposition of deep learning features and traditional image features. \textit{Our proposed framework combines high and low frequency information, visual saliency information, deep learning features and illuminance information of the original image}. The \textbf{illumination information} is used as the non-linear fusion factor of the image fusion to simulate the non-linear fusion characteristics of the human visual system. The \textbf{attention network} is used to simulate the human eye's selection characteristics of fusion features. The \textbf{main contributions} of our work include the following three points:
\begin{itemize}
\item \textbf{Firstly}, in the cross-modal image fusion task, we propose an image fusion method with feature selection characteristics.

\item \textbf{Secondly}, based on the non-linear characteristics of human visual fusion perception, we propose a non-linear cross-modal image fusion method combining illumination factors.

\item \textbf{Finally}, based on the human visual perception mechanism, we propose a robust cross-modal image fusion framework with traditional methods and deep learning knowledge.
\end{itemize}
The remainder of this paper is structured as follows. Sect. 2 reviews relevant theory knowledge. Sect. 3 presents the non-linear and selective fusion of cross-modal images. Sect. 4 introduces the experimental datasets, evaluation metrics, and implementation details. Sect. 5 presents a discussion and explanation. Sect. 6 gets a conclusion.

\section{Related work}
Our research content includes the non-linear fusion characteristics of the human vision system, feature selection characteristics and cross-modal image fusion, so this section will review the existing work from these three aspects.

\subsection{Non-linear fusion characteristic}
The characteristics of visual masking, brightness and contrast sensitivity in the human visual perception system indicate that human perception of external information depends more on the brightness difference between object and background. Human beings have some self-adaptive brightness adjustment function in highlight areas, and human eyes cannot detect the distortion below just noticeable distortion (JND) \cite{Treisman1980A}. However, \textit{the process of adaptive brightness adjustment is one of the non-linear characteristics of the human visual system}. In addition, \textit{human brain as a highly complex non-linear system} \cite{Klonowski2008Importance}, \textit{its processing of information is not simple weighted average, but involves more non-linear processing} \cite{Akay2000Nonlinear}. Inspired by these characteristics, we propose an innovative cross-modal image fusion framework. The fusion rules of the basic level and the detail level of the image will no longer be based on the traditional method (weighted average, maximum, and sum et al.) \cite{Ma2018Infrared}, but on the illumination factor as a non-linear fusion factor. By introducing a non-linear fusion factor, the non-linear fusion characteristics of human visual perceptron are simulated.

\subsection{Feature selection characteristic}
Zohary et al. \cite{Zohary1992Population} pointed out that \textit{physiological evidence shows that visual cortex cells are selective in several perceptual dimensions at the same time, which enables people to select features}. At the same time, Kubovy et al. indicated that \textit{the brain will select specific "features" of the stimulus according to the object of interest, such as direction, spatial frequency or the moving direction of the brightness edge, etc}. Inspired by this, researchers have made relevant achievements in many computer vision fields. Hu et al. \cite{hu2017squeezeandexcitation} proposed a channel attention network for feature selection. Zhang et al. \cite{Zhang_2018RCAN} performed a performance evaluation of channel attention module for residual network on image super-resolution study. Jun et al. \cite{Fu2018Dual} proposed  a dual attention network for image segmentation, and introduced spatial attention module based on channel attention. All the above methods are based on the characteristics of the human visual perception system. The inherent deduction mechanism of vision in the human visual perception system points out that the human visual system deduces content according to prior knowledge in human brain, and discards uncertain information. Inspired by this feature, we use channel attention network \cite{Zhang_2018RCAN} to simulate the feature selection characteristics of the human visual perception system in the cross-modal image fusion tasks. \textit{Attention network is used to learn the complex non-mutually exclusive non-linear relationship between different features, and different weight coefficients are given to the features with different degrees of attention}.

\subsection{Image fusion}
Here we classify the cross-modal image fusion algorithm into traditional image fusion algorithm and deep learning method. We will review the representative algorithms in these fusion algorithms. 

\begin{enumerate}[(i)]
\item \textit{Deep learning image fusion review}. Liu et al. \cite{Liu2016ImageCSR} presented a multi-modal image fusion method based on sparse convolution representation, which overcomes two shortcomings of sparse representation method including insufficient image high frequency feature extraction and very high sensitive to misregistered images. In addition, Liu et al. \cite{Liu2017MultiCNN} proposed a deep convolution neural network framework for multi-focus image fusion. This method is not universal and only suitable for multi-focus image fusion. Li et al. \cite{Li2018DenseFuse} proposed a dense fuse network for infrared and visible images, which uses dense blocks and has multiple hop connections, making full use of the underlying information of the image. Fang et al. \cite{fang2019crossmodal} proposed an image fusion method based on multi-task assistant characteristics for the first time. At the same time, Fang et al. \cite{fang2019crossmodal} presented a cross-modal image fusion method based on subjective visual Attention. There are also some evaluation and research on the application of generative adversarial network in image fusion tasks. Ma et al. \cite{MaFusionGAN} introduced the generative adversarial network into infrared and visible image fusion for the first time. However, the image fusion effect of this method is fuzzy and smooth, lacking of rich texture information, so the Ma et al. \cite{MA202085} proposed a detail preserving learning method. The detail information of fusion image is effectively preserved by detail loss and edge loss \cite{fang2019crossmodal}. In view of the multi-resolution fusion problem of cross-modal data, Ma et al. \cite{9031751} proposed dual-discriminator conditional generative adversarial network based on the network of confrontation generation \cite{fang2019crossmodal}. 

\item \textit{Traditional image fusion review}. 
Based on multi-scale decomposition theory, Li et al. \cite{Shutao2013ImageGF} evaluated pixel saliency and image spatial continuity when calculating the weight of different images for fusion. Bavirisetti DP et al. \cite{Bavirisetti2016Two} proposed an image fusion algorithm combining multi-scale decomposition and visual saliency. Bavirisetti et al. \cite{Zhou2016Fusion} proposed a night-vision context enhancement method to improve the image fusion effect in the dark. To better preserve the texture information, Li et al. \cite{li2018infraredlrr} evaluated latent low-rank representation for image fusion task. Li et al. \cite{Li_2018DL} proposed a pre-training deep convolution neural network based on two-scale decomposition for infrared and visible images. At the same time, Lahoud et al. \cite{Lahoud2019FastZERO} proposed a zero-learning image fusion method, which combines the traditional image features with the deep learning features. This method does not require a specially designed neural network for training. 
\end{enumerate}
\textit{The existing image fusion algorithms have the following problems}. \textbf{Firstly}, the effective feature extraction of images is insufficient. Most algorithms only fuse features directly after extracting them, and there is no secondary selection of features \cite{Liu2017MultiCNN,Liu2016ImageCSR,Li_2018DL,Lahoud2019FastZERO,Li2018DenseFuse}. \textbf{Secondly}, the image fusion method is simple. Most algorithms extract the features directly by simple weighted average, select the weight maximum or extract the feature principal component (PCA) \cite{Ma2018Infrared}. And the non-linear relationship between features is not fully considered, which is not in line with the human visual perception mechanism. \textbf{Finally}, there is still a great gap between the results of cross-modal image fusion of mainstream algorithms and those of human visual fusion. In order to overcome the above problems, we propose a non-linear and selective fusion of cross-modal images. Our method improves the quality of image fusion by simulating human visual characteristics. In our image fusion algorithm, based on the selection characteristics of human visual perception characteristics, the \textbf{attention module is introduced to select the fused image features}. Inspired by the characteristics of human visual perception brightness and contrast sensitivity, we use image \textbf{illumination information to simulate the non-linear combination characteristics} of human eyes with different features in our framework, and establish the non-linear relationship between image fusion based on the illuminance information of the image. \textit{Through the simulation of human visual perception characteristics, the image fusion quality is more in line with human subjective evaluation}.

\section{Method}

\begin{figure}[ht]
	\setlength{\belowcaptionskip}{-0.5cm}
	\centering
	\includegraphics[width=1\textwidth]{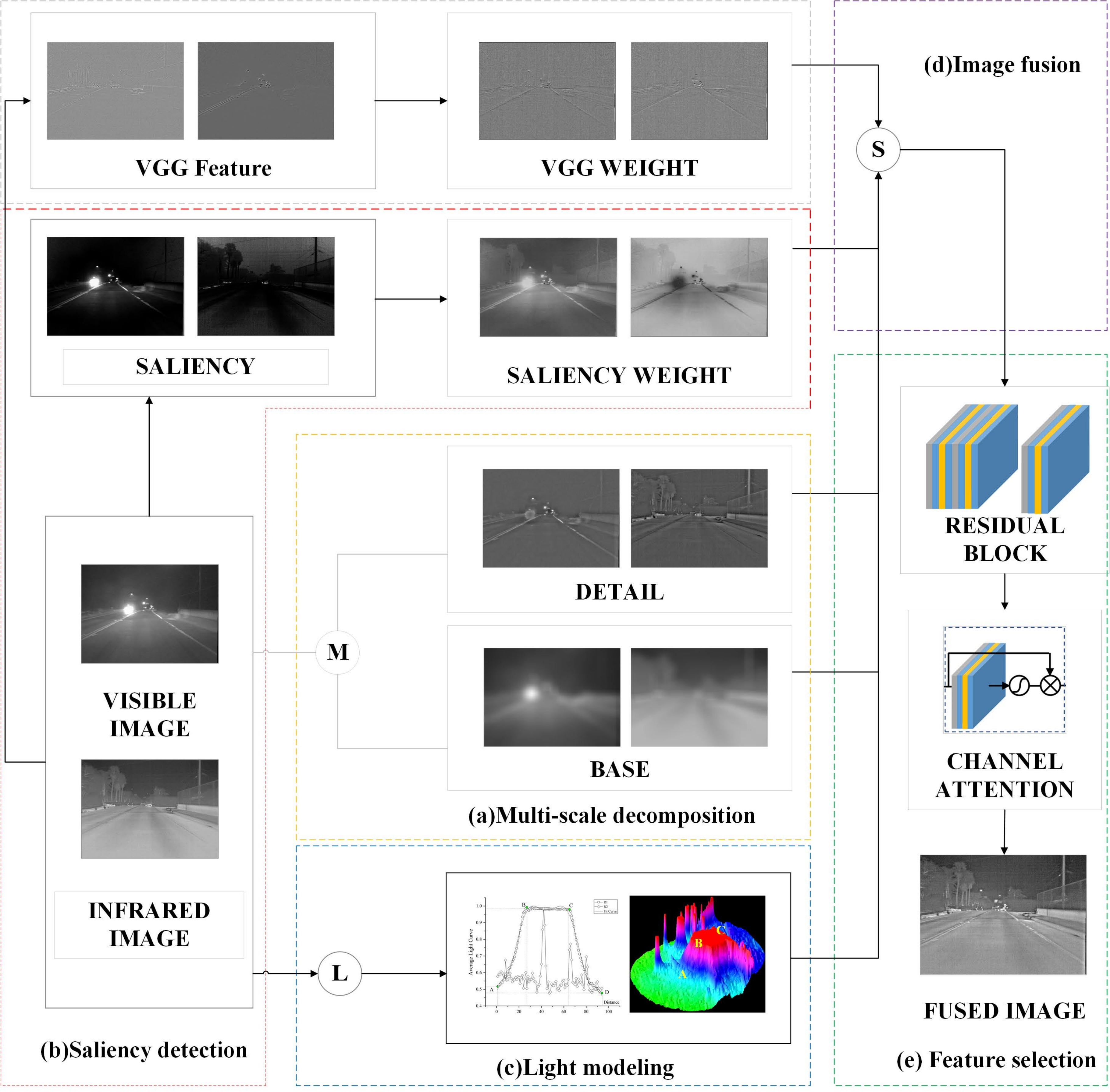}

	\caption{ General block diagram of our framework. Yellow box represents multiscale decomposition of image. Red box indicates significance detection module. The blue box indicates the lighting modeling module. Gray box representation deep feature weight extraction module. Green box representation feature selection module. M, S, L indicate the multi-scale decomposition operation, non-linear fusion functions, highlight block detection respectively. 
}
	\label{general}
\end{figure}

As shown in Figure \ref{general}, our proposed image fusion algorithm needs to complete the following four steps. \textbf{Firstly}, image decomposition is performed to obtain the image base layer and detail layer. \textbf{Secondly}, the image illuminance is modeled, and the non-linear fusion coefficient of the image fusion is obtained. \textbf{Thirdly}, the obtained weight map is combined with the illuminance information fusion factor for feature fusion. \textbf{Finally}, the fused feature map is selected by the channel attention module to obtain the final fused image.

\subsection{Multi-scale image decomposition}
Multi-scale image decomposition theory has been widely used in the field of computer vision, and has achieved great results in feature extraction. According to human visual perception theory, human eyes have different sensitivity to different regions of degraded images. Therefore, in the cross-modal image fusion task, we need to decompose the image at different levels. The method can effectively avoid the image ringing effect caused by high and low frequency mixing during image processing. In our image fusion framework, we use the two-scale decomposition method proposed by \cite{Bavirisetti2016Two}. This method has better real-time performance than the existing multi-scale decomposition methods.

\subsection{Visual saliency detection}
With human's further study of their own visual perception mechanism, saliency detection method based on human visual perception theory has been widely used in the field of computer vision \cite{Li_2018DL}. In image fusion task, the bottom-up and top-down saliency models are usually used, which are realized by the high contrast of the pixels compared with the surrounding information. At present, cross-modal image fusion methods based on saliency detection are mainly two ways, one is to calculate the saliency weight map corresponding to the original image \cite{Lahoud2019FastZERO,Bavirisetti2016Two}, the other is to extract saliency object \cite{Liu2015MultiDSIFT} based on saliency analysis. In this paper, we mainly adopt the bottom-up saliency model method proposed by \cite{Bavirisetti2016Two}, which has less computational complexity than other algorithms. What we need to explain here is that the saliency detection method we use is not to detect object, but to detect the brightness, contrast, edge and other image attributes.

\subsection{Illumination factor modeling}
\begin{figure}[ht]
	\centering
	\includegraphics[width=1\textwidth]{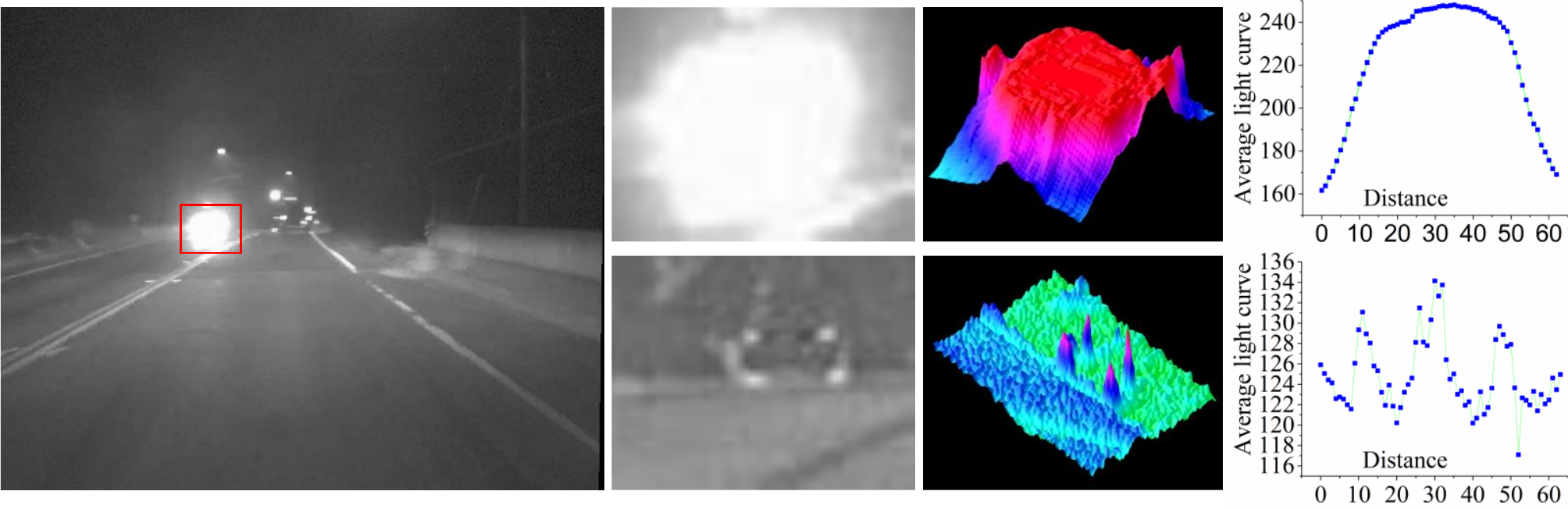}
	\caption{ Visualization analysis of high optical density images. From left to right indicate visible image, abnormal high light image block, abnormal $3D$ optical density map, abnormal average optical density curve $R_1(x, y)$, normal image block, normal $3D$ optical density map, normal average optical density curve $R_2(x, y)$.}
	\label{light}
\end{figure}

In the process of image imaging, the image quality is degraded due to the influence of weather, light and motion. Image quality degradation is due to the loss of high frequency information in the image, and the image information loss part is often presented in the form of high light and dark light. The following figure shows the problem of image information loss caused by car lights in visible images. As shown in Figure \ref{light}, we conducted a visual modeling analysis of the light field.

From the average optical density curve, we can easily find that high light exists in the form of a parabolic cross section. Through a lot of validation on FLIR dataset, it is found that the problem of high light caused by circular light source is universal. However, considering the diversity of light sources (rectangle, ellipse and irregular shape) in nature, we cannot use a fixed illumination model for illumination modeling when doing image highlight removal. We need to establish an illuminance model based on image illuminance, and dynamically adjust the image according to different illuminance models. Aiming at the image fusion task, we validate the current mainstream image fusion algorithm, and the effect is shown in the figure above. From Figure \ref{summary} , we can clearly find that the current mainstream image fusion algorithms do not consider the image highlight problem. Therefore, highlight blocks cannot be effectively removed in the effect of image fusion, and there is obvious boundary effect, which seriously affects the quality of image fusion. Therefore, we propose to introduce illumination factors in image fusion to effectively eliminate the influence of highlights on image fusion quality.

Based on the theory of physical optics, the light perceived by the human eye is mainly composed of ambient light, diffuse reflection light, specular reflection light, and reflection of the object's own light source\cite{Treisman1980A}. To simplify the model, we believe that the image is composed of two parts including the incident image and the reflected image. At present, the incident image estimation method is mainly based on the image low frequency theory, and it is considered that the image illumination is slow, and the image is mainly a low frequency component. Therefore, in various computer vision tasks, the illumination image is generally estimated based on this theory. However, this method has an obvious drawback in that it ignores the non-smooth nature of the illumination itself, which is especially prominent at the edge of the image illumination. Our method fully considers the above problems, and the specific steps are as follows.
\begin{enumerate}[(i)]
\item we transfer the image to YCrCb color space to obtain the brightness information of the image. On this basis, we use morphological operation to detect the high light region of interest, obtain the region of interest corresponding to the image to be fused by mapping operation.

\item Calculation of average light intensity. The average light intensity $L_j(x,y)$ is defined as:
\setlength\abovedisplayskip{2pt}
\begin{gather}
\begin{split}
L_j(x,y) = exp(\frac{1}{h}\sum_{y=0}^{h}ln(\alpha+P(x,y)))
\end{split},
\end{gather}
Where $L_j(x,y)$ represents the average light intensity corresponding to any x-column pixel; $P(x,y)$ represents (x,y) pixel value; $\alpha$ is a very small constant to prevent a pixel value of 0; $h$ represents the sampling height of pixels in the intensity density function image. The setting of this value determines the accuracy of the intensity density function. \textit{When $h$ equals $1$, it is the intensity density function of a row of pixels in the highlight block image. At this time, the accuracy reaches the maximum value. With the increase of $h$, the fitting accuracy of the intensity density function decreases gradually. In our algorithm, $h$ is set to $1$}.

\item Modeling of light intensity density function $R_j(x,y)$.
It can be expressed as n columns $L_j(x,y)$ by polynomial fitting. $R_j(x,y)$ is defined as:
\setlength\abovedisplayskip{2pt}
\begin{gather}
\begin{split}
R_j(x,y) = \sum_{k=0}^{n}\beta L_j(x,y),i=1,2,3,...,n
\end{split},
\end{gather}
Where $\beta$ represents the polynomial coefficient. Therefore, we can transform the tradictional complex illumination modeling problem into the function fitting problem with reference image. 

\end{enumerate}

\subsection{Image fusion}
\begin{figure}[ht]
	
	
	\centering
	\includegraphics[height=8cm,width=11cm]{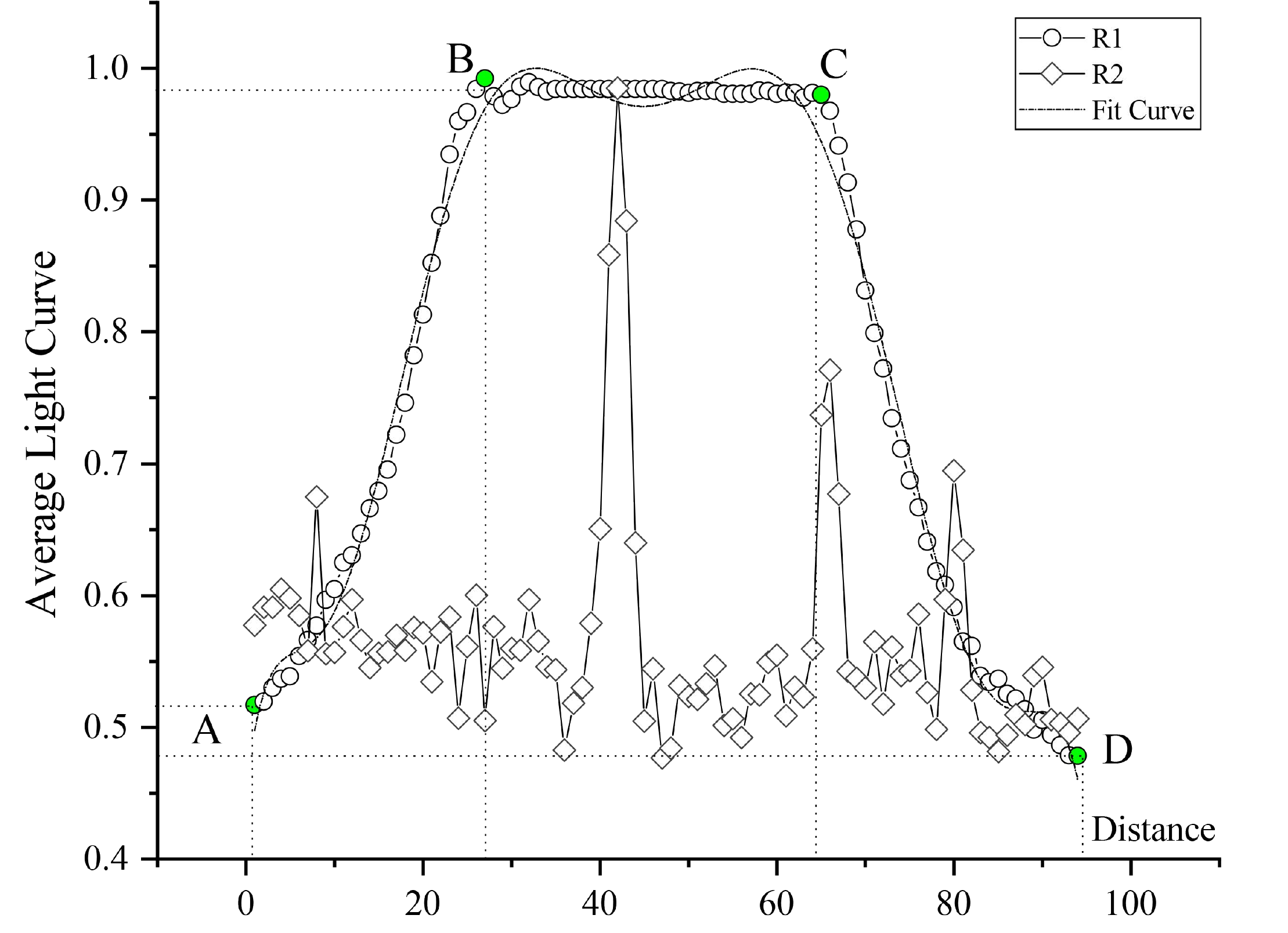}
	\caption{Light intensity curve.}
	\label{Figure11}
\end{figure}

In Figure \ref{Figure11}, $B-C$ segment represents the boundary image of the highlight block image, and the image texture information is seriously lost. $A-B$ and $C-D$ segment represents the transition stage from the highlight image to the surrounding image, and the light intensity curve in this area is non-linear under the influence of the highlight. From $A$ to $B$ and from $D$ to $C$, the fusion weight of the highlight block image gradually decreases. When it reaches the $B-C$ segment, the fusion weight of the highlight block image reaches the minimum value. Therefore, in the process of image fusion, we cannot simply detect the $B-D$ image highlight block for image fusion, otherwise there will be serious fusion boundary phenomenon. In our algorithm, the above problems are effectively overcome by non-linear modeling. On the basis of obtaining the non-linear illumination factor, we calculate the basic level fusion $\overline{B}$, the detail level fusion $\overline{D}$, and the final image fusion $\overline{F}$. The specific definitions are as:

\setlength\abovedisplayskip{2pt}

\begin{equation}
\left\{
\begin{array}{rl}
\overline{B} & =\sum_{i=1}^{n}\left(\frac{W_{i}^{b} * B_{i}}{1+\exp \left(-C * R_{b}(x, y)-\frac{1}{2}\right)}+\left(1-\frac{W_{i+1}^{b} * B_{i+1}}{1+\exp \left(-C * R_{b}(x, y)-\frac{1}{2}\right)}\right) \right), \\
\overline{D} & =\sum_{i=1}^{n}\left(\frac{W_{i}^{d} * D_{i}}{1+\exp \left(-C * R_{d}(x, y)-\frac{1}{2}\right)}+\left(1-\frac{W_{i+1}^{d} * D_{i+1}}{1+\exp \left(-C * R_{d}(x, y)-\frac{1}{2}\right)}\right) \right), \\
\overline{F} & =\overline{B}+\overline{D},
\end{array}
\right.
\end{equation}
Where $W$ denotes saliency weight map, $i$ denotes an $i$-th image; b denotes a base layer superscript; $d$ denotes a detail layer; $x$ and $y$ denote pixel coordinates; $C$ denotes a pixel normalization constant. 

In our image fusion framework, we use normalized illumination intensity as the non-linear illumination coefficient. Considering that the non-linear fusion factor needs to be in the range of $0$ and $1$, we use sigmoid as the activation function. We can find that the weighted average fusion criterion or maximum fusion criterion at the basic level and detail level of the current mainstream algorithms is a special case of our algorithm. When the illumination factor is $0.5$, it is the weighted average algorithm in the current mainstream fusion method, and maximizing the illumination factor is the maximum fusion criterion. 

\subsection{Feature selection}

We suppose that the length, width and channel obtained by residual convolution after previous fusion are $W\times H\times C$. As shown in Equation \ref{gs4} , the global average pooling (GP) operation is performed on the $T$ feature map to obtain the global receptive field corresponding to the feature map, so that the network can exclude the spatial relationship between different channels and focus on learning the non-linear relationship between different feature channels. The output $C_{AM}$ after feature selection is defined as \cite{fang2019crossmodal,Zhang_2018RCAN}:

\setlength\abovedisplayskip{1pt}
\begin{gather}
C_{AM}=S(W_2,R(W_1,F_{GP}))*T_{k},
\label{gs4}
\end{gather}
Where $Tk(x, y)$ represents the pixel value corresponding to the kth channel $(x, y)$ coordinates. After passing through the global average pooling layer, we obtain the output of the attention module through convolution, RELU activation function, convolution, Sigmoid activation function, and dot product operation; $S$ and $R$ represent the activation functions of Sigmoid and Relu respectively, while $W_1$ and $W_2$ represent the weight of two convolutions respectively; $F_{GP}$ indicates the output of the input image after GP operation. 

\section{Experiments}
\label{setup}
In this section, experimental setup are presented and comparative experiments result produced along with relevant explanations and discussions are presented.

\subsection{Experimental setup}
In this section, datasets, metrics and methods for experimental evaluation are presented. Finally, implementation details of evaluated methods are introduced.
\subsubsection{Datasets}

	\textit{1) TNO} \cite{TNO}: It has five pairs of infrared and visible images. It contains multi-spectral (enhanced vision, near infrared and long wave infrared or thermal) night images of different military related scenes, registered in different multi-band camnera systems. There are 21 pairs of image pairs commonly used in existing image fusion algorithms.
	
	\textit{2) FLIR} \cite{FLIR}: A set of annotated thermal images and non annotated RGB images is provided for the training and verification of the neural network for target detection. The data set is obtained by RGB and thermal imaging camera installed on the vehicle. The dataset contains 14452 annotated hot images, of which 10228 are from short video and 4224 are from 144 second video. Unfortunately, there is no registration.
	
	\textit{3) ATLAS} \cite{Summers2003Harvard}: It includes 97 CT and MRI images and 24 T1-T2 weighted MRI images. The relevant images have registered the data.
	
	\textit{4) VIFB} \cite{zhang2020vifb}: Vifb is the first (and only) benchmark in the field of visible infrared image fusion (VIF). It aims to provide a fair and comprehensive performance comparison platform for Vif algorithm. At present, vifb integrates 21 image pairs, 20 fusion algorithms and 13 evaluation indexes, which can be used for performance comparison. Fortunately, 20 algorithms corresponding to 21 images provide fused images. Unfortunately, no specific code has been released.

\begin{table}[!h]
	
	\centering \footnotesize
	\renewcommand \arraystretch{1.1}
	\caption{
	Experimental datasets and inherited properties
}
	\label{table11}
	\begin{tabular}[b]{p{1.4cm}p{1.6cm}<{\centering}p{2cm}<{\centering}p{2.2cm}<{\centering}p{1cm}<{\centering}p{1cm}<{\centering}}
		\specialrule{0.08em}{5pt}{5pt}
		Dataset & Scene & Challenge & Modality & Registration & Matching pairs\\ \hline
		TNO \cite{TNO}  &  Military   & Illumination, noise  &  Infrared and visible  &  \checkmark&63\\ 
		FLIR \cite{FLIR} & Highway & Illumination, noise   & Infrared and visible  &$\times$& 14452 \\ 
		ATLAS \cite{Summers2003Harvard} & Medical & Noise, imaging    & CT, MRI  &\checkmark&97    \\ 
		VIFB \cite{zhang2020vifb} & City & Illumination  & Infrared and visible  &  \checkmark&21  \\ 
		\specialrule{0.08em}{2pt}{0pt}
	\end{tabular}
\end{table}

The main properties of experimental datasets are summarized in Table. \ref{table11}. To evaluate the robustness of our framework, we performed experimental evaluations on different image fusion task data sets. 

\subsubsection{Metrics}
The distinctiveness of an image quality is usually quantitatively evaluated using entropy (EN) \cite{1576816}, average gradient (AG) \cite{Cui2015Detail}, structural similarity (SSIM) \cite{1284395}, mutural information (MI) \cite{Qu2002Information}, visual information fidelity (VIF) \cite{Han2013A}, information fidelity criterion (IFC) \cite{Sheikh2006An}. 

\begin{enumerate}[(i)]

\item \textit{EN} \cite{1576816} represents information entropy. Information theory points out that the higher the information entropy is, the better the image quality is. EN is defined as:
\begin{equation}\mathrm{EN}=-\sum_{i=0}^{255} p_{i} \log _{2} p_{i},\end{equation}
Where $P_i$ is the probability of a gray level appearing in the image, which can be obtained by gray histogram.

\item \textit{AG} \cite{Cui2015Detail} represents average gradient. It reflects the change rate of small detail contrast and represents the relative clarity of the image. Generally speaking, the higher the evaluation gradient, the higher the image level. AG is defined as:
\begin{equation}A G=\frac{1}{M^{*} N} \sum_{i=1}^{M} \sum_{j=1}^{N} \sqrt{\frac{\Delta I_{x}^{2}(i, j)+\Delta I_{y}^{2}(i, j)}{2}},\end{equation}
Where $M \times N$ denotes the image height and width; $\Delta I_{x}(i, j)$ denotes image horizontal gradient; $\Delta I_{y}(i, j)$ denotes image vertical gradient.

\item \textit{SSIM} \cite{1284395} denotes structureal similarity. The image quality is evaluated from three aspects: brightness, contrast and structure. The mean value is used as the estimation of brightness, the standard deviation as the estimation of contrast, and the covariance as the measurement of structural similarity. SSIM is defined as:
\begin{equation}\begin{array}{l}
\operatorname{SSIM}(I_i, R)=\frac{\left(2 u_{I_i} u_{R}+C_{1}\right)\left(2 \sigma_{I_i R}+C_{2}\right)}{\left(u_{I_i}^{2}+u_{R}^{2}+C_{1}\right)\left(\sigma_{I_i}^{2}+\sigma_{R}^{2}+C_{2}\right)}, \\
\text { SSIM }=(\operatorname{SSIM}_{I_1, R}+\operatorname{SSIM}_{I_2, R})/2.0,
\end{array}\end{equation}

Where $\mu_{I_i}$ and $\mu_R$ indicate the mean value of origin image $I_i$ and fused image $R$; $\sigma_{I_iR}$ is the standard covariance correlation.

\item \textit{MI} \cite{Qu2002Information} denotes mutual information. It indicates the correlation between two images. The more similar the images are, the greater the mutual information is. MI is defined as:

\begin{equation}M I(I_i, R)=H(I_i)+H(R)-H(I_i, R),\end{equation}
Where $H(I_i)$ and $H(R)$ represent the information entropy of origin image and fused image; $H(I_i,R)$ denotes joint information entropy.

\item \textit{IFC} \cite{Sheikh2006An} represents information fidelity Criterion. The image quality is evaluated by measuring the amount of common information between the original image and the fused image. IFC is defined as:

\begin{equation}\mathrm{IFC}\approx \alpha(\mathrm{PDC})+N_{\mathrm{sub}} \beta,\end{equation}
Where $PDC$ denotes perceptual distortion criterion; $k$ denotes the index of the k-th sub-band, and $N_{sub}$is the
number of subbands used in the computation; $\alpha$ and $\beta $ are constants.

\item \textit{VIF} \cite{Han2013A} represents visual Information Fidelity. The image quality is evaluated by simulating the significant physiological and psychological visual characteristics of the human visual system (HVS). The larger the value is, the better the image quality is, and the more consistent with the human visual perception system. VIF is defined as:

\begin{equation}\mathrm{VIF}=\frac{\sum_{j \in \text { subbands }} I\left(\vec{C} \stackrel{N, j}{;} \vec{F}^{N, j} | s^{N, j}\right)}{\sum_{j \in \text { subbands }} I\left(\vec{C}^{N, j} ; \vec{E}^{N, j} | s^{N, j}\right)},\end{equation}
Where $\vec{C} \stackrel{N, j}{;}$ denotes N elements of the $C_j$ that describes the coefficients from subband j; $\sum_{j \in \text { subbands }} I\left(\vec{C} \stackrel{N, j}{;} \vec{F}^{N, j} | s^{N, j}\right)$ denotes reference image information.

\end{enumerate}

\subsubsection{Methods}

\begin{table}[!h]
	
	\centering \footnotesize
	\renewcommand \arraystretch{1.1}
	\caption{PARAMETER SETTINGS OF EVALUATED METHODS}
	\label{table12}
	\begin{tabular}[b]{p{0.05cm}p{1.8cm}<{\centering}p{3.5cm}<{\centering}p{0.3cm}<{\centering}p{1.3cm}<{\centering}p{0.3cm}<{\centering}p{0.7cm}<{\centering}p{0.7cm}<{\centering}}
		\specialrule{0.08em}{5pt}{5pt}
		No.& Method &Parameters &Year  & Category & Light & Selection &Time(s) \\ \hline
		1&FEZ \cite{Lahoud2019FastZERO}&$r_b = 35, \varepsilon_b = 0.01, r_d = 7, \varepsilon_d = 1e-6$&2019&Hybrid& $\times$&$\times$&0.56 \\ 
		2&CSR \cite{Liu2016ImageCSR}&$\lambda=0.01$ &2016&Multi-scale& $\times$&$\times$&98.14 \\ 
		3&DL \cite{Li_2018DL}& $\alpha_1=\alpha_2=0.5, k\subset[1,2]$&2018& Deep learning & $\times$&$\times$ &18.62\\ 
		4& DENSE \cite{Li2018DenseFuse}&$Epoach=4, Lr=0.0001$ &2019& Deep learing& $\times$&$\times$&0.83\\ 
		5&FusionGAN \cite{Ma2018Infrared}&$Epoach=10, Lr=0.0001$ &2019&GAN& $\times$&$\times$&0.10\\ 
		6&IFCNN \cite{ZHANG202099}&$L_{r0}=0.01, power=0.9$&2020&Deep learning & $\times$&$\times$&0.08 \\
		7&DTCWT \cite{Lewis2007Pixel}&$\times$&2007& Wavelets & $\times$&$\times$&0.25\\ 
		8&LATLRR \cite{Li2018InfraredLTLRR}&$\lambda =0.4, stride=1$ &2020&Multi-scale  & $\times$&$\times$&271.04\\ 
		9&LP-SR \cite{Liu2015ALPSR}&$overlap=6, \epsilon =0.1, level=4$&2015&Hybrid& $\times$&$\times$&0.04\\ 
		10&DSIFT \cite{Liu2015MultiDSIFT}&$Scale=48, blocksize=8, matching=1$&2015&Other& $\times$&$\times$&3.98 \\ 
		11&CNN \cite{Liu2017InfraredCNN} & $t=0.6$ &2017&Hybrid& $\times$&$\times$&31.76 \\ 
		12&CVT \cite{Nencini2007RemoteCVT} &$is_real=1, finest=1$&2007&Multi-scale & $\times$&$\times$&1.09\\ 
		13&CBF \cite{Shreyamsha2015ImageCBF} &$\sigma s=1.8, \sigma r=25, ksize=11$&2015&Multi-scale& $\times$&$\times$&22.97\\ 
		14&JSR \cite{Zhang2013Dictionary}&$Unit=7, step=1, dic_size=256, k=16$	&2013&Sparse representation& $\times$&$\times$ &93.89\\ 
		15&JSRSD \cite{Liu2017InfraredJSR-SD} &$Unit=7, step=1, dic_size=256, k=16$&2017&Saliency-based& $\times$&$\times$&172.44 \\ 
		16&GTF \cite{Ma2016InfraredGTF}&$Epsr=epsf=tol=1, loops=5$ &2016&Other & $\times$&$\times$&6.27 \\ 
		17&WLS \cite{Ma2017InfraredWLS}&$\sigma_s=2, \sigma_r=0.05, nLevel=4$&2017&Hybrid& $\times$&$\times$&8.18\\ 
		18&RP \cite{Toet1989ImageRP}&$\times$&1989&Pyramid& $\times$&$\times$&0.76 \\ 
		19&MSVD \cite{Naidu2011Image}  &$\times$&2011&Multi-scale& $\times$&$\times$&1.06\\ 
		20&MGFF \cite{Durga2019Multi}&$R=9, \varepsilon=10^3, k=4$&2019&Multi-scale  & $\times$&$\times$&1.08\\
		21 &ZCA \cite{Li2018Infrared}& $K=2, i=4 and i=5$  &2019&Hybrid&$\times$&$\times$&2.57 \\
		22&ADF \cite{Bavirisetti2016Fusion} &$w1=w2=0.5$&$2016$&Multi-scale & $\checkmark$&$\times$&1.00\\
		23&FPDE \cite{Bavirisetti2017Multi} &$At=0.9, n=20, k=4, \delta t=0.9$&$2017$&Subspace & $\checkmark$&$\times$&2.72\\
		24&IFEVIP \cite{Zhang2017Infrared} &$Nd = 512, Md = 32, Gs = 9, MaxRatio = 0.001, StdRatio = 0.8 $&$2017$&Other & $\checkmark$&$\times$&0.17\\	
		25&MGFF \cite{Durga2019Multi} &$\times$&$2019$&Multi-scale & $\checkmark$&$\times$&1.08\\	
		26&OURS &$h=1$&$2020$&Hybrid & $\checkmark$&$\times$&0.76\\
		27&OURS+ &$h=1$&$2020$&Hybrid & $\checkmark$&$\checkmark$&1.09\\
		\specialrule{0.08em}{2pt}{0pt}
	\end{tabular}
\end{table}

As shown in Table \ref{table12}, we will compare experiments with 25 mainstream algorithms such as fast-zero-learning (FEZ) \cite{Lahoud2019FastZERO}, fonvolutional sparse representation (CSR) \cite{Liu2016ImageCSR}, deep learning (DL) \cite{Li_2018DL}, dense fuse (DENSE) \cite{Li2018DenseFuse}, generative adversarial network for image fusion (Fusion GAN) \cite{Ma2018Infrared}, laplacian pyramid (LP) \cite{Burt1987TheLP}, dual-tree complex wavelet transform (DTCWT) \cite{Lewis2007Pixel}, latent low-rank representation
(LATLRR) \cite{Li2018InfraredLTLRR}, multi-scale transform and sparse representation (LP-SR) \cite{Liu2015ALPSR}, dense sift
(DSIFT) \cite{Liu2015MultiDSIFT}, convolutional neural network (CNN) \cite{Liu2017InfraredJSR-SD}, curvelet transformation (CVT) \cite{Nencini2007RemoteCVT}, bilateral filter fusion method (CBF) \cite{Shreyamsha2015ImageCBF}, cross joint sparse representation (JSR) \cite{Zhang2013Dictionary}, joint sparse representation with saliency detection (JSRSD) \cite{Liu2017InfraredJSR-SD}, gradient transfer fusion (GTF) \cite{Ma2016InfraredGTF}, weighted least square optimization (WLS) \cite{Ma2017InfraredWLS}, a ratio of low pass pyramid (RP) \cite{Toet1989ImageRP}, multi-resolution singular value decomposition (MSVD) \cite{Naidu2011Image}, non-linear (OURS), non-linear fusion and feature selection (OURS+). At the same time, we need to point out that the time efficiency of different algorithms is tested on the VIFB data set. As the fourth data set VIFB gives the image fusion results of related algorithms, some of these algorithms are repeated with the algorithms we have compared, and other algorithms will be introduced in Section \ref{VIFB}. In Table 2, the time is calculated using the average time obtained from the VIFB data set.

\subsubsection{Implementation details}
Before the experiment, we need to clarify the following questions.
\textit{1)} In all subsequent experiments, we converted all images into grayscale images for subsequent image fusion. \textit{2)} We also need to point out that the robustness problem in this paper is mainly verified from two aspects. On the one hand, it is verified from multiple cross-modal datasets. On the other hand, it is verified from complex environment, such as high light and dark light images. Therefore, this paper will not test the robustness separately, which will be reflected in each experiment. \textit{3)} For different experiments, there will be some changes in the related algorithm experiments, and the changes will be explained in the respective experimental chapters. 
\textit{4)} These algorithms have already published their code, and the relevant algorithm parameters are the same according to the settings in the public paper. 
\textit{5)} For our proposed algorithm, we also conducted a comparative experiment on whether there is a channel attention module or not. 
\textit{6)} Aiming at the problem that FLIR data set is not registered, we use matlab toolbox to do some manual alignment work.
\textit{7)} Since VIFB is a new image fusion benchmark proposed in 2020, but the number and time of the algorithms contained in the benchmark are not dominant. Therefore, in the first three data sets, we adopt the latest image fusion algorithms and some classic image fusion algorithms. Of course, the most important thing is that the code base is not public.
\textit{8)} Although the VIFB dataset does not provide a code base, so we only tested nine of them, and only showed six of them commonly used in Figure \ref{f15}. \textit{9)} Our experimental platform is desktop 3.0 GHZ i5-8500, RTX2070, 32G memory.

\begin{figure}[ht]
	
	
	\includegraphics[scale=0.8,width=1\textwidth]{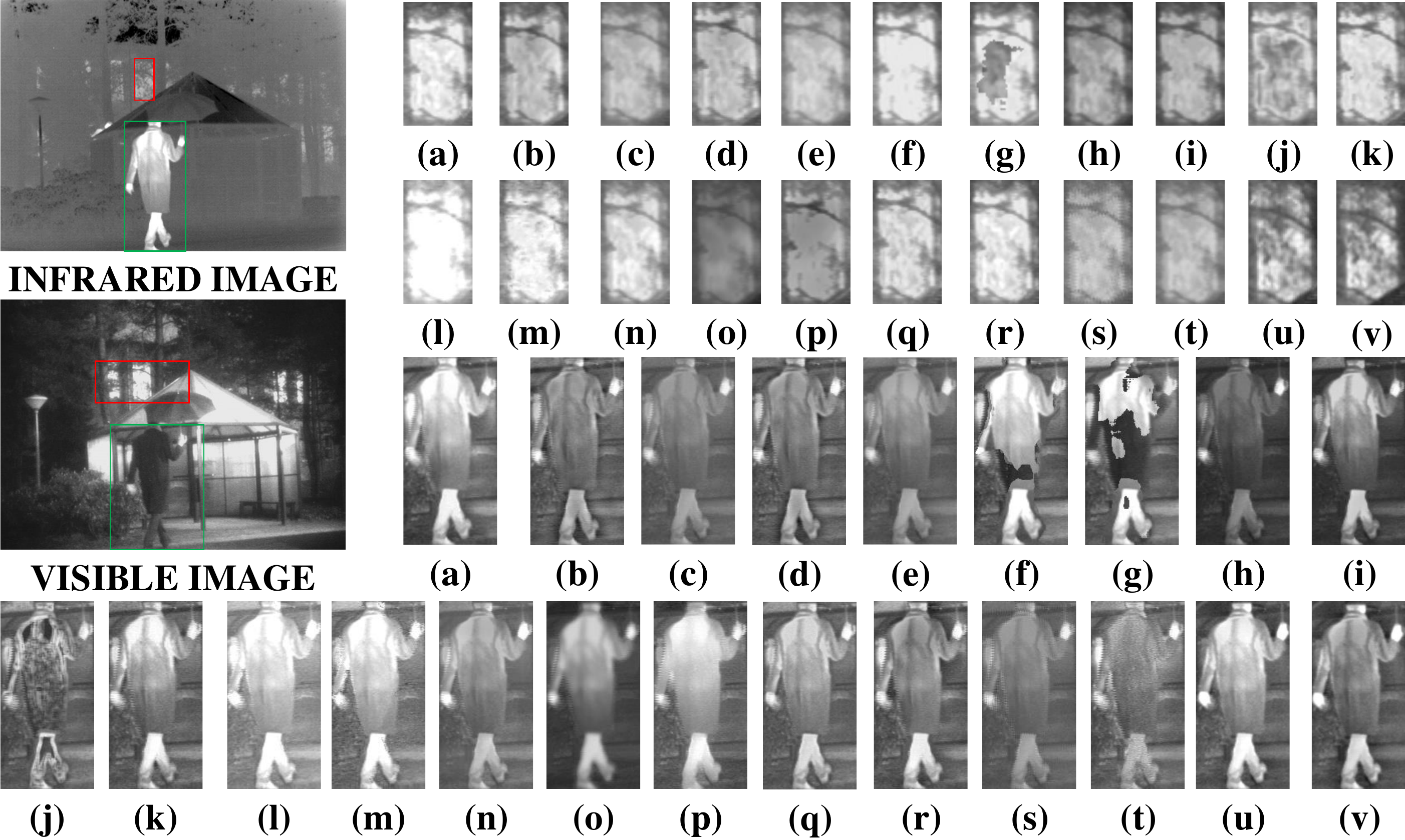}
	\centering
	
	\caption{
		Visible and infrared source images with the fusion results obtained by different methods. From (a) to (v) : CNN\cite{Liu2017InfraredCNN}, CVT\cite{Nencini2007RemoteCVT}, DL\cite{Li_2018DL}, DTCWT\cite{Liu2015MultiDSIFT}, FEZ\cite{Lahoud2019FastZERO}, DSIFT\cite{Liu2015MultiDSIFT}, CSR\cite{Liu2016ImageCSR}, DFA\cite{Li2018DenseFuse}, DFL1\cite{Li2018DenseFuse}, CBF\cite{Shreyamsha2015ImageCBF},
		WLS\cite{Ma2017InfraredWLS},
		JSR\cite{Zhang2013Dictionary}, JSRSD\cite{Liu2017InfraredJSR-SD}, LATLRR\cite{Liu2011Latent}, FusionGan\cite{MaFusionGAN}, GTF\cite{Ma2016InfraredGTF}, IFCNN\cite{ZHANG202099}, LPSR\cite{Liu2015ALPSR}, MSVD\cite{Naidu2011Image}, RP\cite{Toet1989ImageRP}, 
		OURS, OURS+.}
	\label{f5}
\end{figure}

\subsection{Comparative experiments}
In this section, in order to verify the existing problems of image fusion algorithm and the robustness of our algorithm, we will carry out comparative experiments and visual display on TNO data set, FLIR data set, medical data set and VIFB data set. In subsection \ref{4.2.5}, we will verify the effectiveness of feature selection \cite{ZhaoT.2019Pfan}. In subsection \ref{4.2.6}, we will verify analysis experiment of non-linear fusion.

\subsubsection{Results on TNO dataset}
On TNO \cite{TNO} dataset, we performed quantitative and qualitative analysis on 21 pairs of infrared and visible images in the dataset using the 20 image fusion methods shown in Section \ref{setup}. As shown in Figure \ref{f5}, we have qualitatively analyzed the data set. From the tree leaf window in the above figure, we can see that there is obvious highlight in the visible image, and the image information is seriously lost, but in the infrared image, the detailed structure information of this place is relatively well preserved. Existing algorithms do not have good image restoration to recover lost information in visible images. Compared to other algorithms, our algorithm has a very high definition in the highlights of the trees in the highlights. In the pedestrian window, we can also see that our algorithm allows pedestrians to maintain high contrast information. The images of our algorithm fusion are more in line with the human visual perception mechanisms.

\begin{figure}[!ht]
	
	
	\centering
	\includegraphics[scale=0.8,width=1\textwidth]{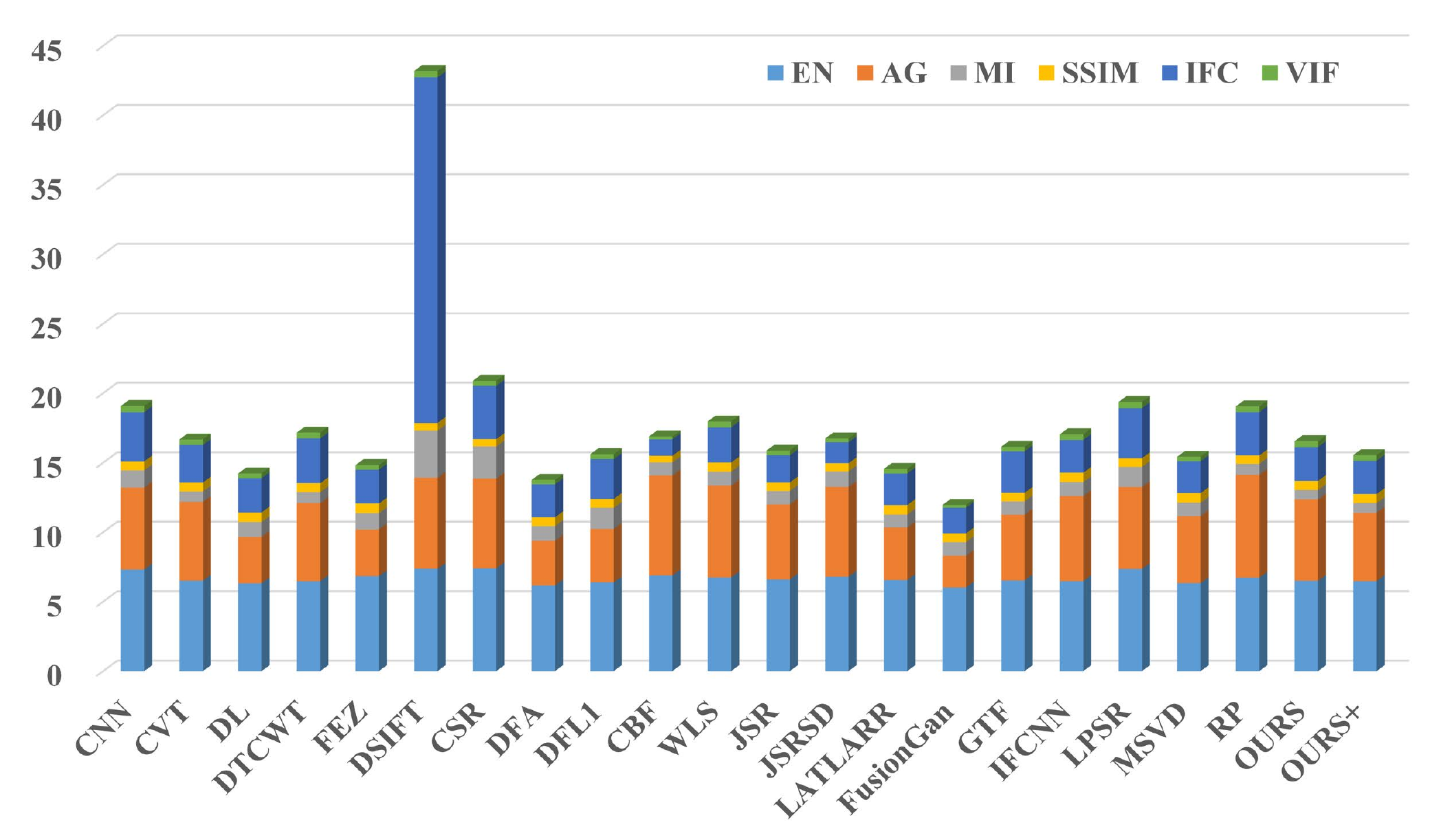}
	\caption{ Six evaluation indicators for quantitative contrast between infrared and visible Images.}
	\label{f6}
	
\end{figure}

Based on the qualitative analysis of the TNO dataset, we performed a quantitative analysis of the dataset. From Figure \ref{f6}, we can find that CSR algorithm and DSIFT algorithm have great advantages in EN, AG, MI and IFC indexes, but when we look at the fusion effect image, we can find that the subjective effect of these two images is the worst, and there are a lot of fusion boundary effects. These objective indicators will mislead image quality assessment. This shows a problem that the existing objective image quality evaluation indicators have their own limitations, and can not be more perfect evaluation of image quality. Although our algorithm does not have an advantage in the objective index in this dataset, our algorithm can effectively avoid the boundary effect while fully preserving the original details of the image.

\subsubsection{Results on FLIR dataset}
\begin{figure}[!ht]
	
	
	\centering
	\includegraphics[width=1\textwidth]{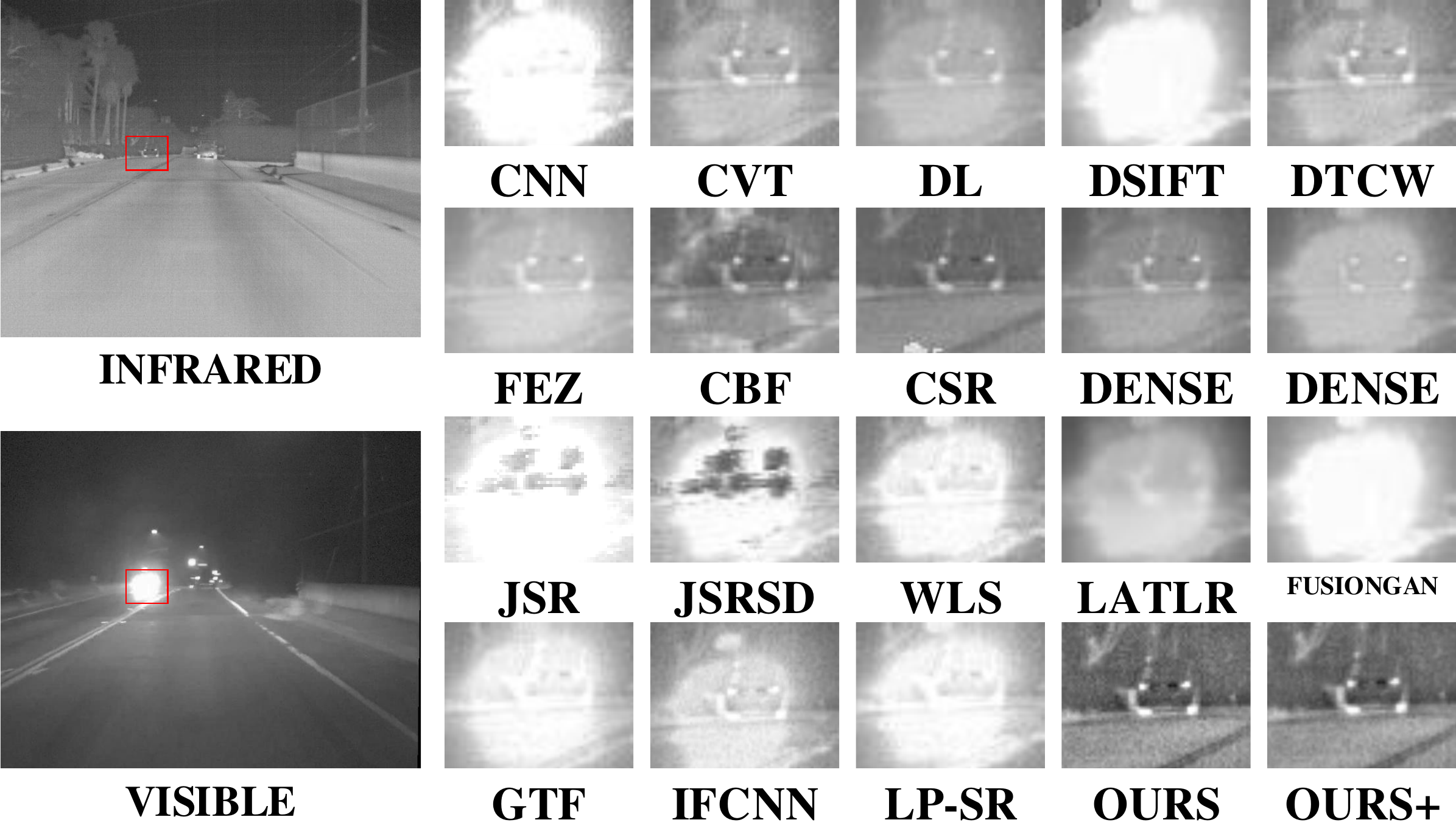}
	\caption{ Qualitative fusion results on visible and thermal infrared images by different method.}
	\label{f7}
\end{figure}

\begin{figure}[!ht]
	
	
	\centering
	\includegraphics[width=1\textwidth]{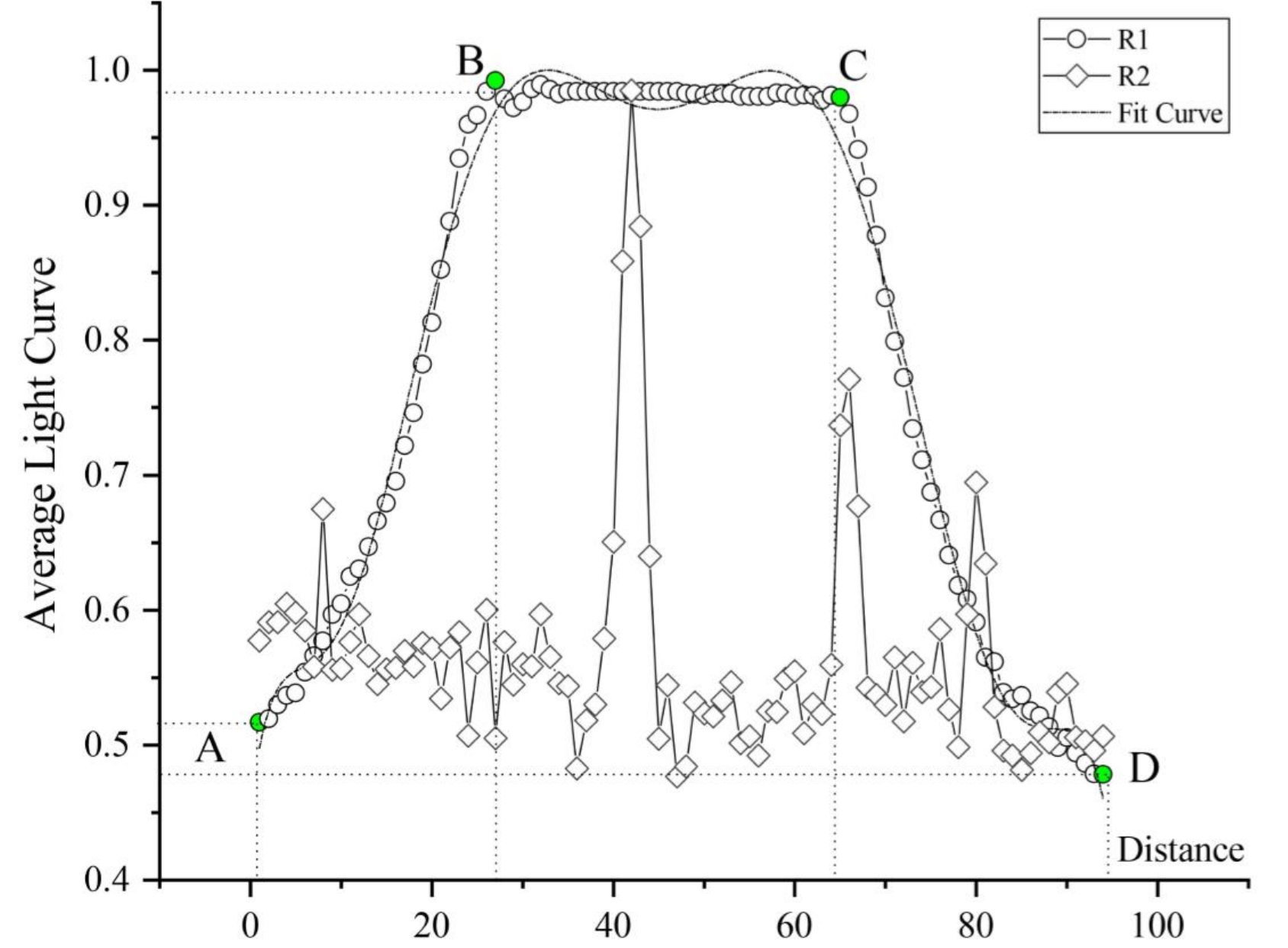}
	\caption{ Six evaluation indicators for quantitative contrast between thermal infrared and visible images.}
	\label{f8}
\end{figure}

In order to verify the robustness of our method, we also carried out related experiments on the FLIR \cite{FLIR} traffic dataset. Subjective visual analysis is shown in Figure \ref{f7} , and the relevant quantitative analysis is shown in Figure \ref{f8} .

From the Figure \ref{f7} , we can see that in the FLIR traffic dataset, our algorithm has higher image fusion quality than other algorithms in the high light block. Our algorithm can effectively remove the highlight and avoid the boundary effect of image fusion. In FLIR data set, DSIFT and CSR are still the worst subjective effects, but the objective indicators are very high. In addition to the boundary effect, the two algorithms seriously lose the detailed texture information of the visible image. At the same time, we observe the SSIM evaluation index and the highlight block image. We can find that the image texture details of the fusion of CVT \cite{Nencini2007RemoteCVT}, DTCWT \cite{Liu2015MultiDSIFT} and RP \cite{Toet1989ImageRP} algorithm have not been repaired at all. we can find that in the FLIR dataset, the SSIM evaluation indexes of these three algorithms are generally more than one percentage point compared with the proposed algorithm. At the same time, these three algorithms have higher visual fidelity than other algorithms. \textit{The reason for the analysis is mainly due to the influence of the brightness and contrast characteristics of the human visual system and the visual masking characteristics. When the image is seriously degraded, the SSIM evaluation index is significantly different from the subjective evaluation. Therefore, when the image is degraded seriously, it is not the higher the SSIM value, the better the image quality.}

\subsubsection{Results on medical dataset}

\begin{figure}[!ht]
	
	
	\centering
	\includegraphics[width=1\textwidth]{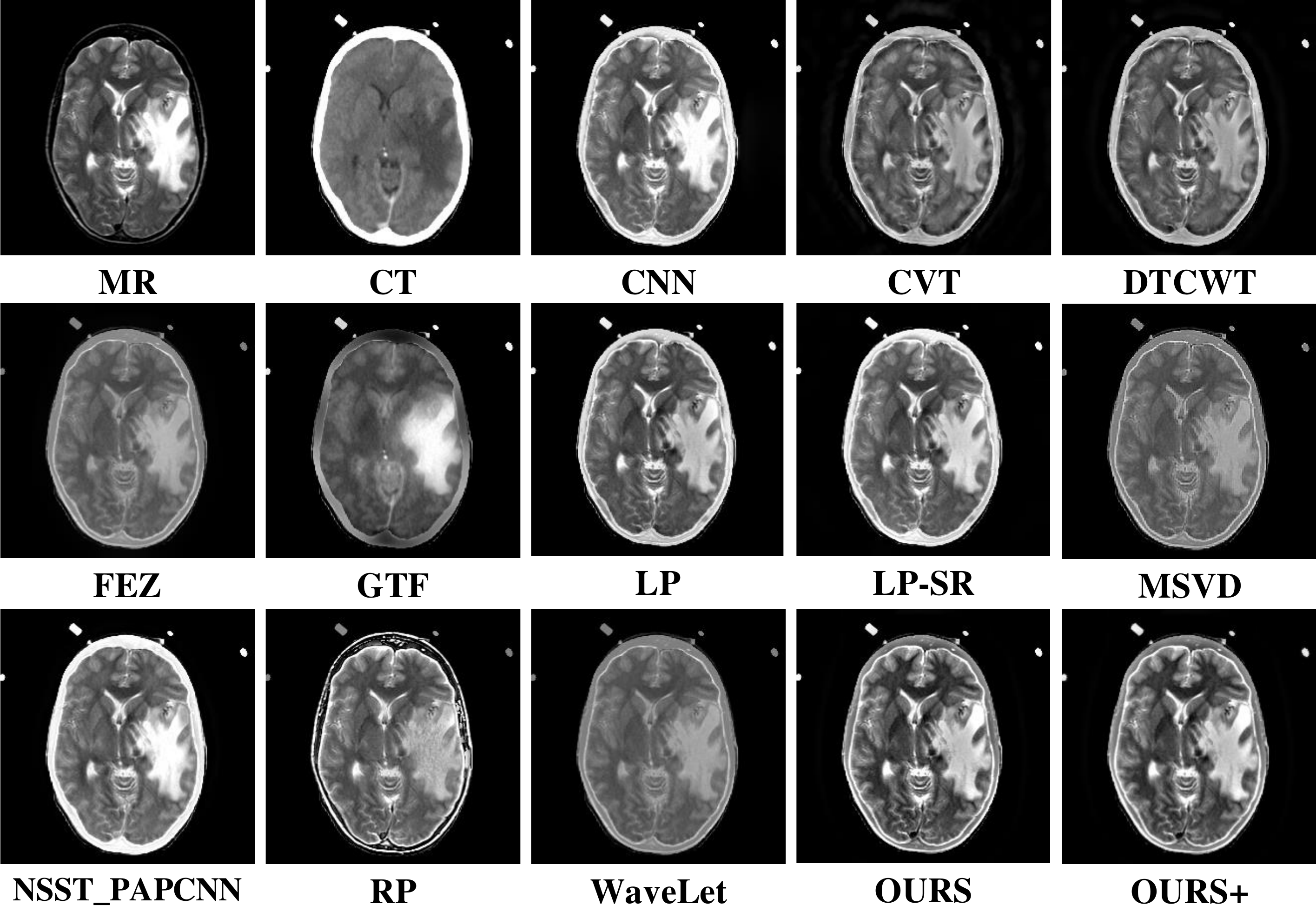}
	\caption{ Qualitative fusion results on CT and MR images by different method.}
	\label{f9}
\end{figure}

\begin{figure}[!ht]
	
	
	\centering
	\includegraphics[width=1\textwidth]{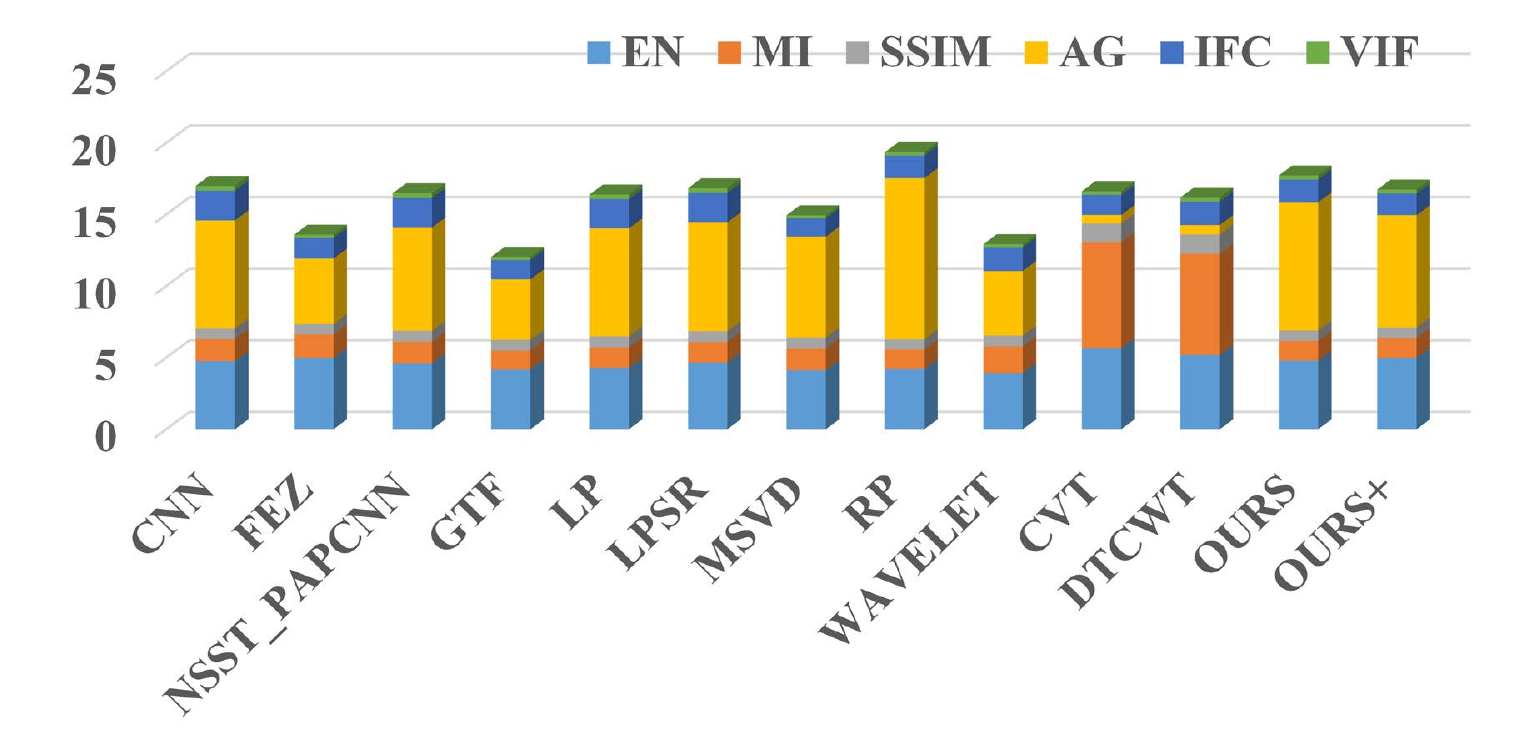}
	\caption{ Qualitative fusion results on CT and MR images by different method.}
	\label{f16}
\end{figure}

In order to further demonstrate the robustness of our algorithm, we tested on the medical image data set, as shown in Figure 9 and Figure 10. Through comparison, we can find that our image fusion effect and CNN algorithm have better clarity. Of course, we are not the best in objective indicators, mainly because the image fusion effect of RP algorithm has obvious fragmentation effect , resulting in a very high gradient value.

\subsubsection{Results on VIFB benchmark}
\label{VIFB}
\begin{figure}[!ht]
	
	
	\centering
	\includegraphics[width=1\textwidth]{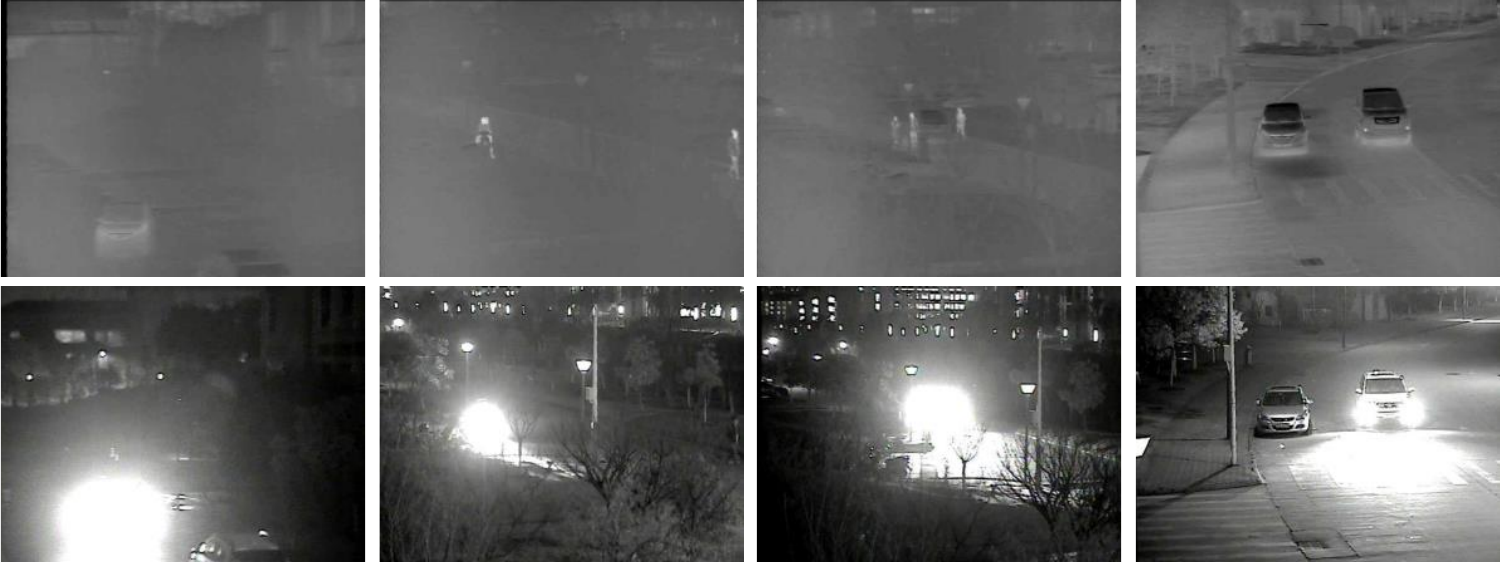}
	\caption{Exemplar infrared and visible images from the VIFB datasets.}
	\label{f10}
\end{figure}

\begin{figure}[!ht]
	\centering
	\includegraphics[width=1\textwidth]{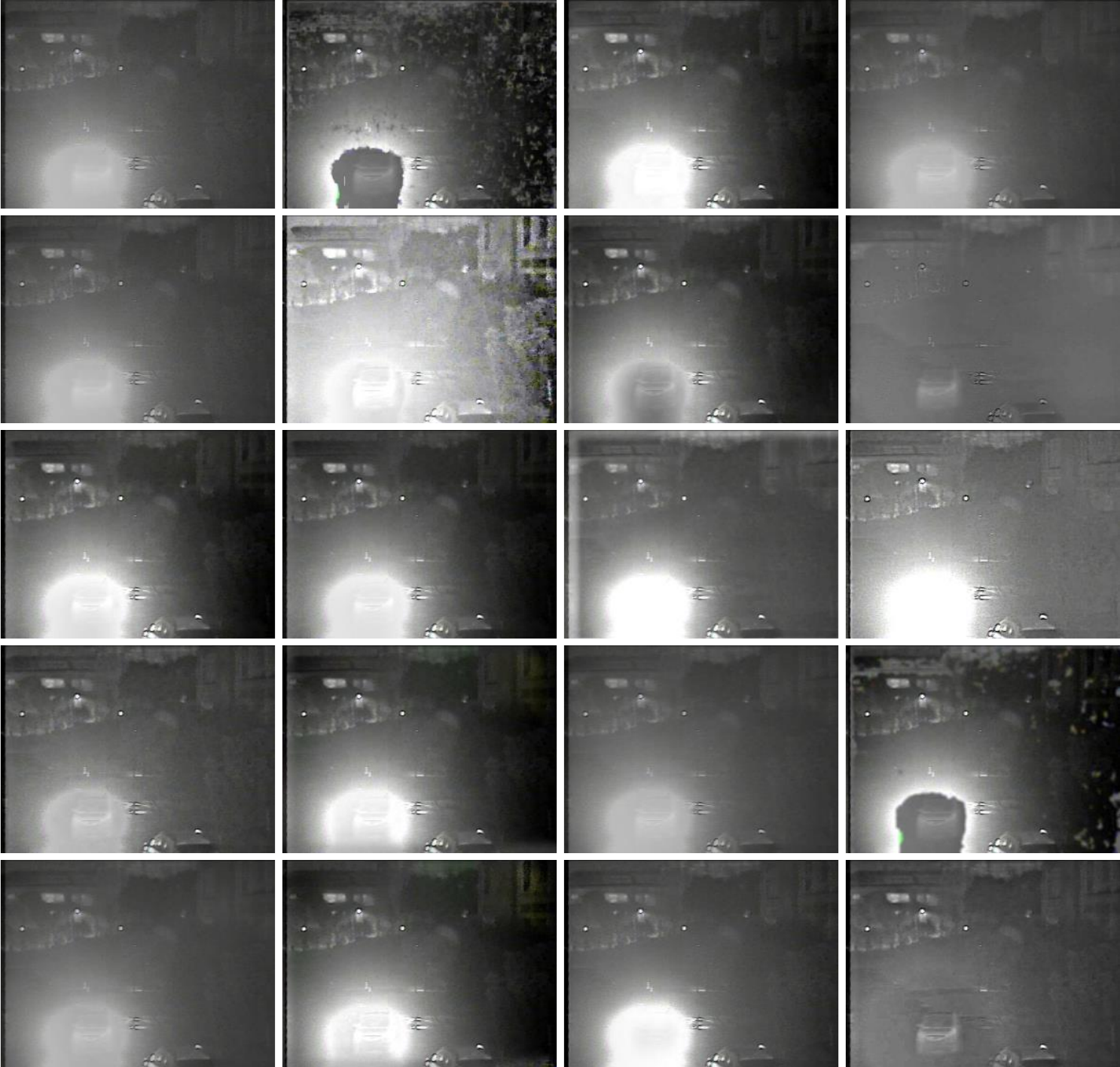}
	\caption{ Qualitative fusion results of carLight images on VIFB dataset.}
	\label{f11}
\end{figure}

\begin{figure}[!ht]
	
	
	\centering
	\includegraphics[width=1\textwidth]{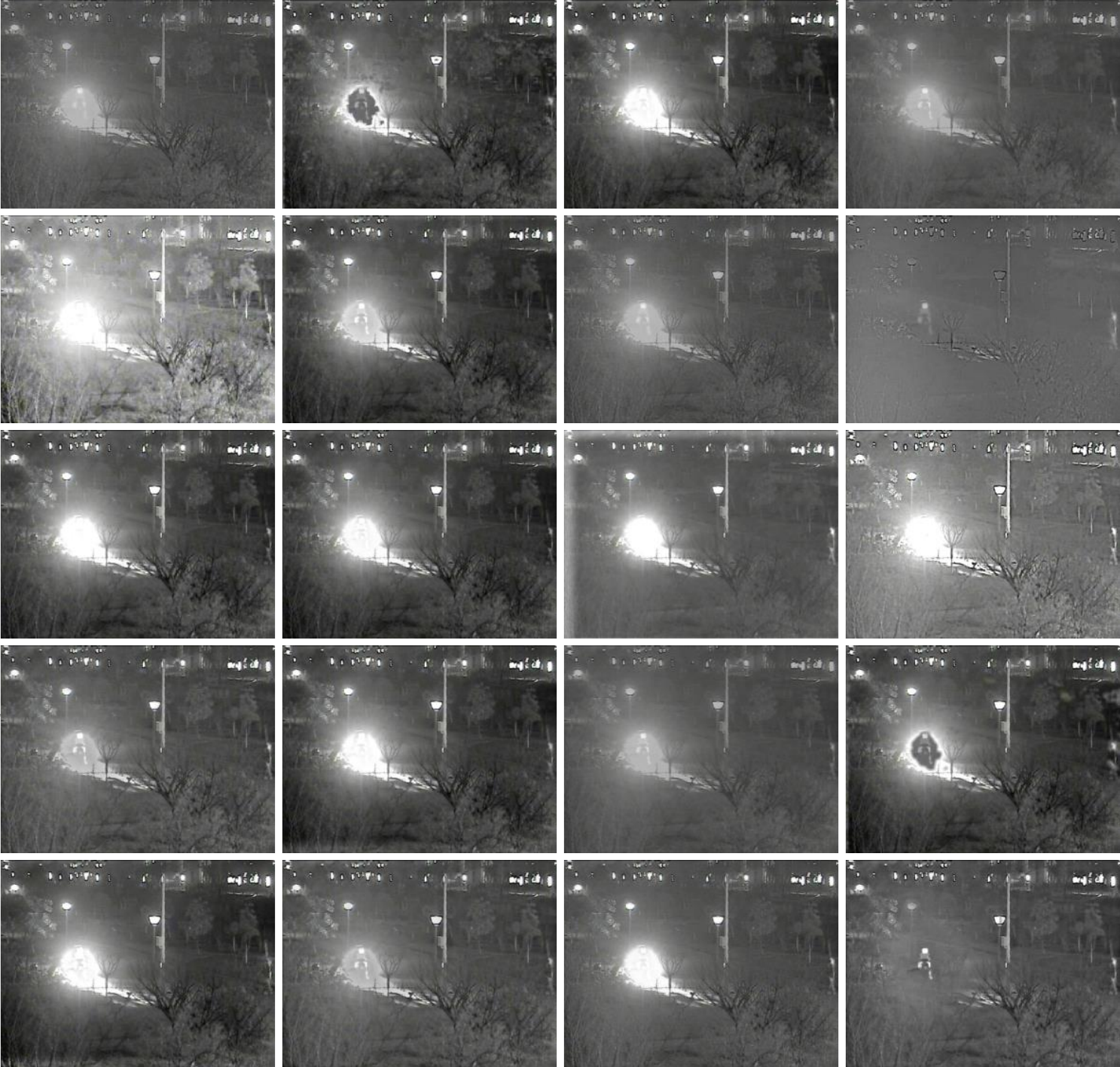}
	\caption{ Qualitative fusion results of elecbike images on VIFB dataset.}
	\label{f12}
\end{figure}

\begin{enumerate}[(i)]
\item{In order to further verify the robustness of the algorithm, we have carried out experiments on the existing infrared and visible image public dataset, as shown in Figure \ref{f10} , which is the image under the complex environment selected from VIFB dataset. From left to right, "carLight", "elecbike", "manlight", "tricycle". The first line is an infrared image, and the second line is a visible image. Refer to Figure \ref{f11}, Figure \ref{f12}, Figure \ref{f13} and Figure \ref{f14} for image fusion effect. In Figure \ref{f11}, Figure \ref{f12}, Figure \ref{f13} and Figure \ref{f14}, \textit{from left to right, from top to bottom} are anisotropic diffusion (ADF) \cite{Bavirisetti2016Fusion}, bilateral filter fusion method (CBF) \cite{Shreyamsha2015ImageCBF}, convolutional neural network (CNN) \cite{Liu2017InfraredJSR-SD}, deep learning (DL) \cite{Li_2018DL}, fourth order partial differential equations (FPDE) \cite{Bavirisetti2017Multi}, night-vision
context enhancement (GFCE) \cite{Zhou2016Fusion}, image fusion with guided filtering (GFF) \cite{Shutao2013ImageGF}, gradient transfer fusion (GTF) \cite{Ma2016InfraredGTF}, night vision context enhancement (HMSDGF) \cite{Zhou2016Fusion}, hybrid multi-scale decomposition with gaussian and bilateral filters (HybridMSD) \cite{Zhou2015Perceptual}, feature extraction and visual information preservation (IFEVIP) \cite{Zhang2017Infrared}, latent low-rank representation (LATLRR) \cite{Li2018InfraredLTLRR}, multi-scale guided image and video fusion (MGFF) \cite{Durga2019Multi}, multi-scale transform and sparse representation (MSTSR) \cite{Liu2015ALPSR}, multi-resolution singular value decomposition (MSVD) \cite{Naidu2011Image}, NSCTSR \cite{Liu2015ALPSR}, zero-phase component analysis (ZCA) \cite{Li2018Infrared}, RP \cite{Toet1989ImageRP}, weighted least square optimization (WLS) \cite{Ma2017InfraredWLS}. }

\begin{figure}[!ht]
	
	
	\centering
	\includegraphics[width=1\textwidth]{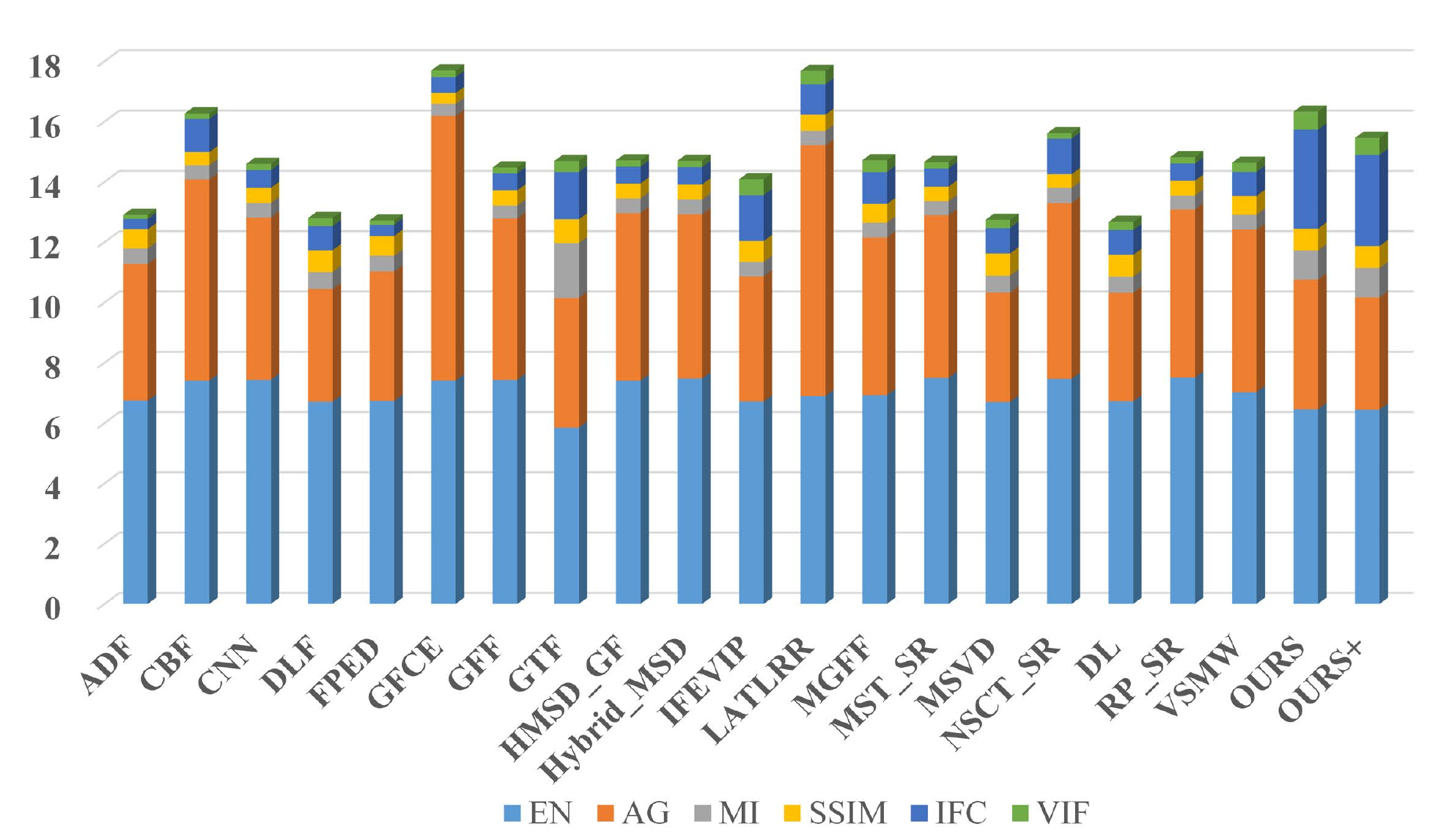}
	\caption{Qualitative fusion results on VIFB dataset.}
	\label{f15}
\end{figure}

\begin{figure}[!ht]
	
	
	\centering
	\includegraphics[width=1\textwidth]{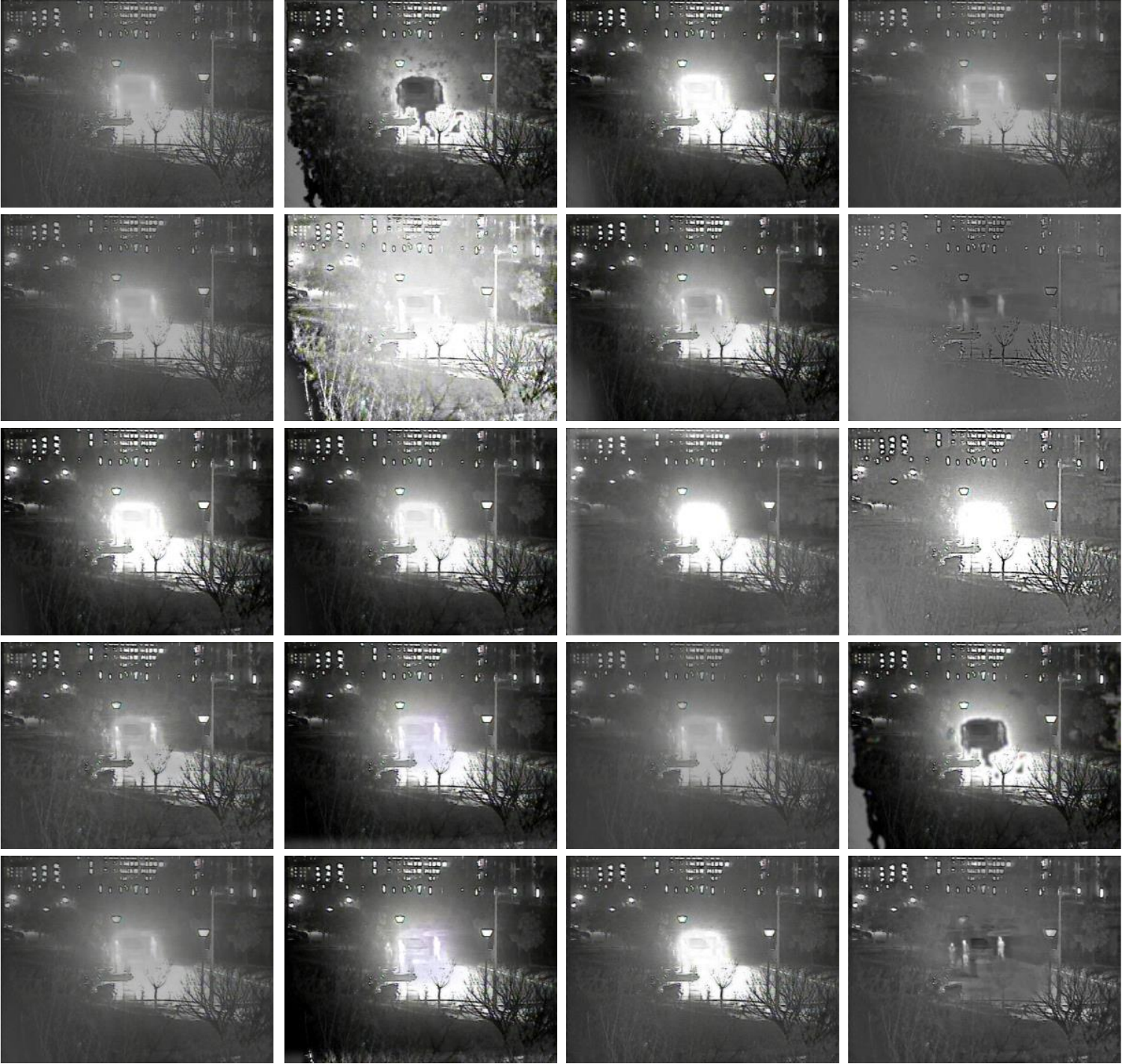}
	\caption{Qualitative fusion results of manlight images on VIFB dataset.}
	\label{f13}
\end{figure}

\begin{figure}[!ht]

	\centering
	\includegraphics[width=1\textwidth]{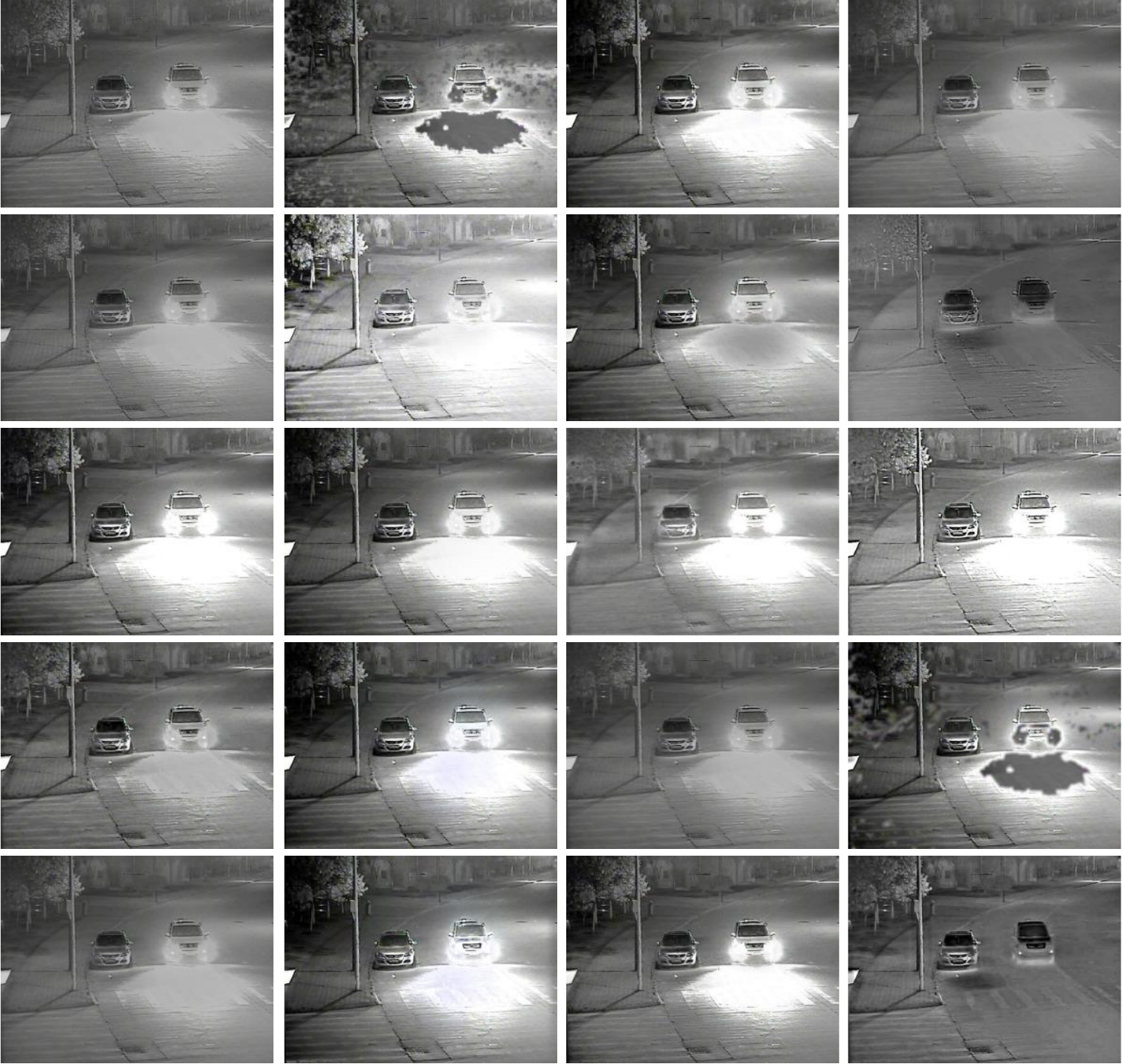}
	\caption{Qualitative fusion results of tricycle images on VIFB dataset.}
	\label{f14}
\end{figure}

\item From Figure \ref{f11}, Figure \ref{f12}, Figure \ref{f13} and Figure \ref{f14}, we can find that our image fusion algorithm has better fusion boundary in the case of high light and dark light. \textit{1)} At the same time, our algorithm fully retains the detailed texture information of infrared image and visible image. \textit{2)} Among all the existing algorithms, the fusion effect of GTF seems to completely remove the highlights and halos, but is this algorithm really effective in image fusion $?$ No. We can see the fusion effect of four groups of test images. It is not difficult to find that the algorithm uses the infrared image as the base image in fusion, but loses a lot of visible image detail texture information, the brightness information and contrast information of the visible image. At the same time, the contrast information of infrared image is also reduced and blurred. Therefore, the fusion effect of the algorithm in complex environment is still not good. \textit{3)} Although our algorithm does not completely remove the high light halo, it has better subjective effect than the existing image fusion algorithm. \textit{4)} From Figure \ref{f15}, we find that although the existing algorithms are not good at image fusion in complex environment, but its average gradient and structural similarity indicators are very high. Such as CBF, GFCE, LATLARR, GTF, et al. \textbf{There are three main reasons}. i) Structural similarity index can not effectively evaluate image quality in complex environment. ii) The structural similarity index belongs to the full reference image quality evaluation index, which lacks the representation ability for the cross-modal image. iii) High gradient usually means better image definition and better image quality, but in these test images, the main reason is that the fused image has obvious edge vibration and noise, rather than the fused image quality is very good. At the same time, from Figure \ref{f14} and Figure \ref{f15}, we find that NSCT\_ SR algorithm has no boundary effect, but its EN and AG are also very high, resulting in the overall index higher than us. However, when we observe the fused image, we find that the algorithm only enhances the visible image, and introduces some low-frequency information of the infrared image, while the high-frequency information of the detail texture of the infrared image is seriously lost. Although our method does not have an advantage in EN and AG evaluation metrics, our image fusion effect is more in line with human subjective vision, which is not only reflected in human subjective feelings, we can clearly find that our index has a greater advantage from the comparison of IFC and VIF evaluation metrics. Of course, from Figure \ref{f12}, we can also find that in the over highlights stage, the texture information of our visible image seems not obvious. This is because in order to prevent the boundary oscillation caused by high frequency information, we also have non-linear fusion of high frequency information.
\end{enumerate}

\subsubsection{Validity analysis experiment of feature selection feature}
\label{4.2.5}
\begin{figure}[!ht]
	
	
	\centering
	\includegraphics[width=1\textwidth]{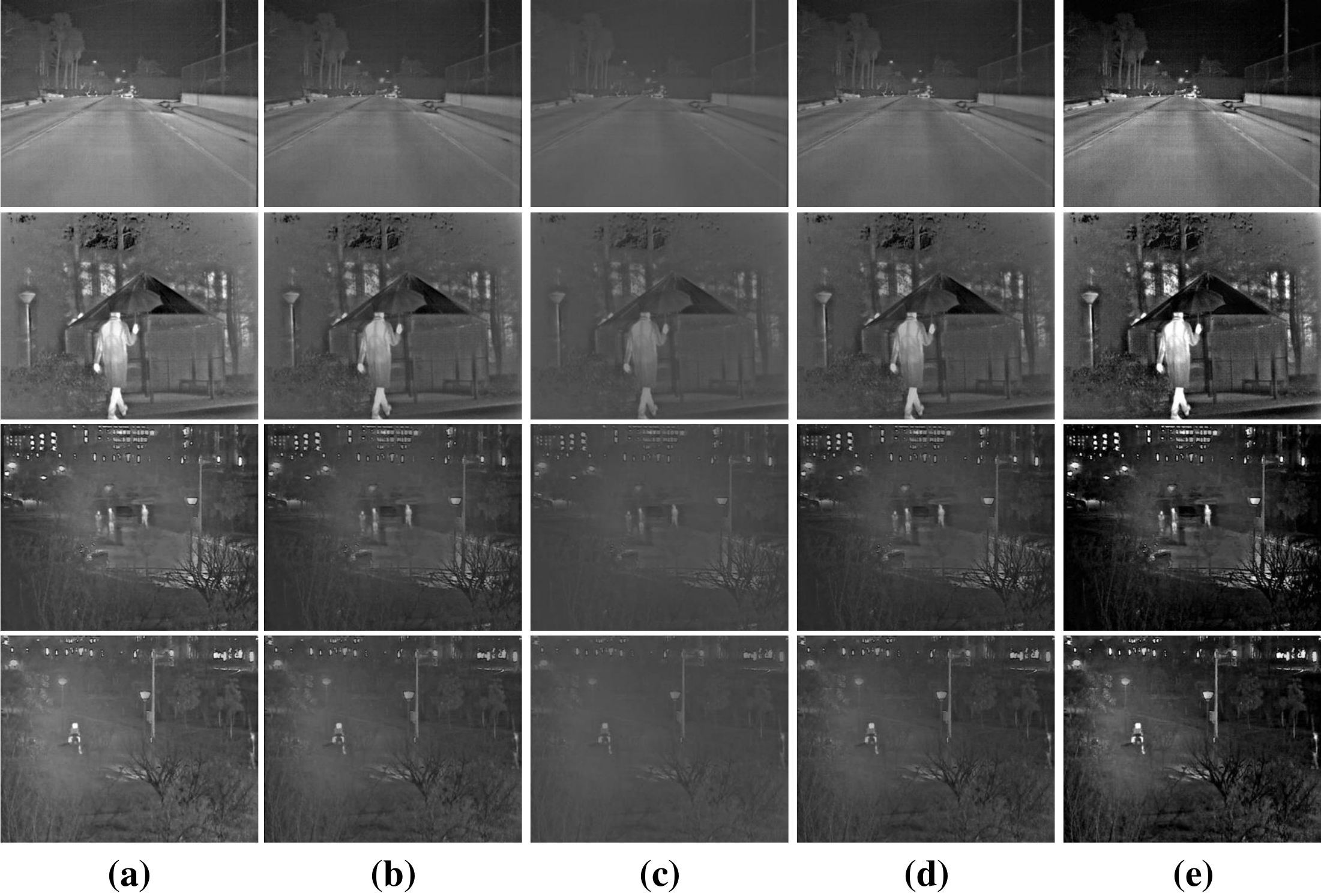}
	\caption{Validity analysis experiment of feature selection feature. (a) On-linear image fusion result. (b) and (c) use the same network weight, the selective attention module was not added in the training, but (c) the attention selection module was added in the test phase. Similarly, (d) and (e) use the same network weight, but add the selective attention module in the training, (e) relatively and (d) cancel the selective attention module in the test phase. (a), (b), (c) and (d) all use the same network parameters and data for training.
	}
	\label{f17}
\end{figure}

Through the comparative experiments on four datasets, we find that adding attention selection module will reduce the overall indicators compared with not adding. The information entropy, average gradient and information fidelity decrease a little, but the mutual information, structure similarity and peak signal-to-noise ratio evaluation index will be slightly improved. To some extent, this proves some conclusions of RCAN \cite{Zhang_2018RCAN} proposed by Zhang et al. It also proves the effectiveness of the feature selection attention feature. According to the different importance of the feature, different weights are given to the feature map. This will inevitably affect information entropy, gradient and information fidelity. However, this kind of influence is hard to detect in subjective vision, so we involve the following experiments to analyze the influence of feature selection characteristics on image fusion visually.

From Figure \ref{f17}, we find that the effect of attention selection on image quality is obvious even if there is no objective quality comparison. Even if we add the attention selection module in the training phase, but cancel the attention module in the test phase, this selection feature also has a great impact on the model weight, it is obvious that the image clarity and contrast have been improved. Compared with the existing algorithm, the effect is significantly improved. But if we do not add attention selection module in the training phase, only use it in the test phase, this way will reduce the quality of the image.  This experiment also proves the effectiveness of introducing feature selection mechanism into the field of image fusion, and to some extent shows that human visual selection feature has a positive effect on image fusion.

\subsubsection{Validity analysis experiment of non-linear fusion}
\label{4.2.6}
\begin{figure}[!ht]
	
	
	\centering
	\includegraphics[width=1\textwidth]{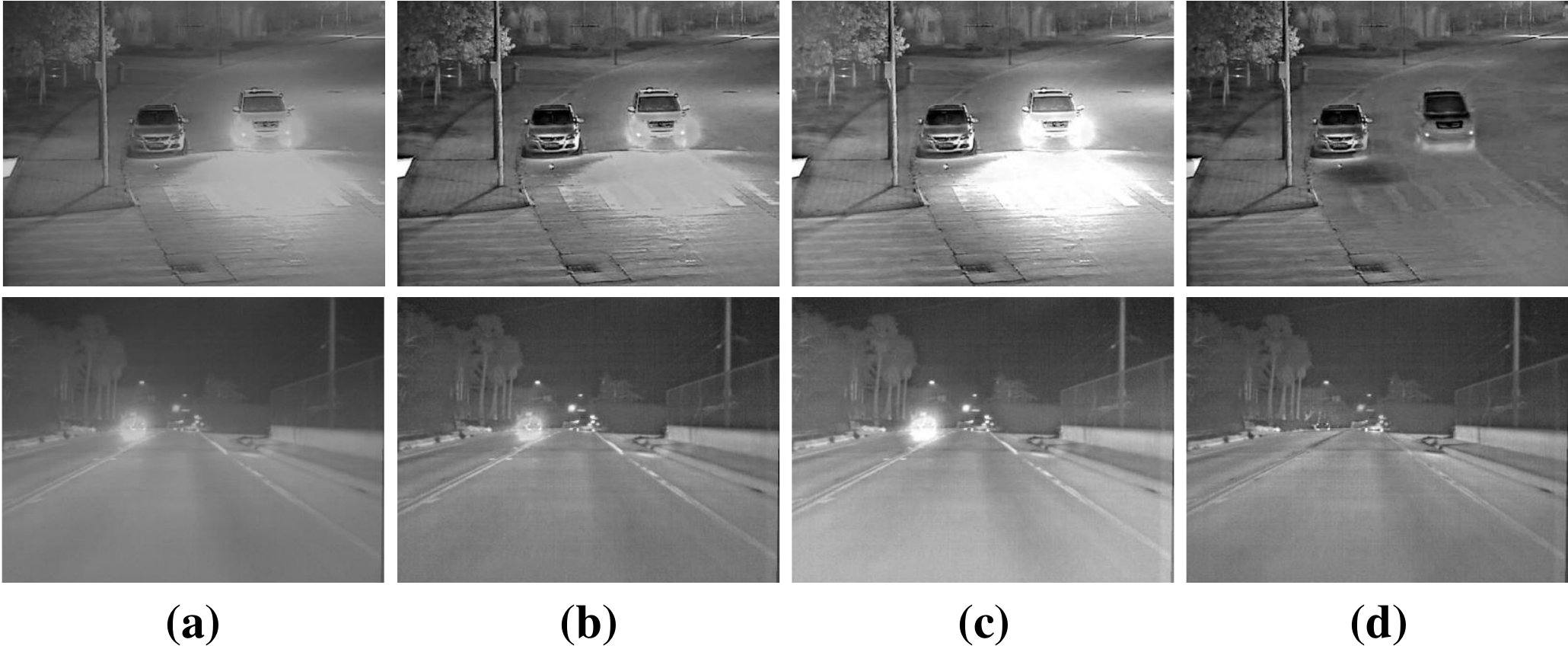}
	\caption{Validity analysis experiment of non-linear fusion. (a) Weighted average image fusion results. (b) Sum image fusion results. (c) Maximum image fusion results. (d) Non-linear image fusion result.}
	\label{f18}
\end{figure}
In order to further verify the advantages of non-linear fusion, we use our algorithm to combine different fusion criteria for experimental comparison. In this experiment, only the fusion criteria are modified, including the maximum fusion criteria, weighted average fusion criteria and sum fusion criteria. From Figure \ref{f18}, we can find that compared with weighted average fusion criterion, sum fusion criterion and maximum fusion criterion, non-linear fusion method has better performance in subjective fusion effect, but this does not mean that its objective index will be very high. Experiments also show the effectiveness of non-linear fusion and the robustness of human visual characteristics in the field of image fusion. Although our method does not fully simulate this non-linear characteristic, the existing results can prove the correctness of this viewpoint to a certain extent.

\section{Discussion}
From the extensive experiments in Section \ref{setup} , it is proved that the proposed image fusion method \textbf{is more in line with human visual system than the existing methods}. We think the main reasons are as follows. \textbf{Firstly}, the collaboration of traditional and deep learning methods is effective in image fusion tasks. \textbf{Secondly}, illumination as a non-linear factor of feature fusion is consistent with human visual perception characteristics. \textbf{Finally}, in the task of image fusion, feature selection is not only effective in the initial stage of feature extraction, but also very important in the later stage of feature fusion. 

In the experiment, \textbf{we also get some interesting phenomenon}. We find that in a complex environment, the image fusion effect of many existing algorithms is not best subjective quality, but the objective quality evaluation index is very high. This illustrates two problems. 

\textit{1)} In the image quality evaluation, the existing papers or benchmark mostly prove the image quality according to the objective evaluation index. Of course, if it is for the image fusion task with ground truth, it is the right way. However, this method is also used for the cross-modal image fusion task without ground truth, which is not so accurate in the experiment. 

\textit{2)} On the other hand, it also points out the limitations of objective indexes in evaluating the quality of cross-modal images.

Although our algorithm has achieved relatively good results in several image fusion tasks, \textbf{there are still some shortcomings}. Human visual perception system is a very complex system. \textit{When processing image fusion tasks, human beings will add more understanding and cognition from pixel level to semantic level, which is incomparable to the existing image quality evaluation function.} When processing image fusion tasks, human beings will add more understanding and cognition from pixel level to semantic level, which is incomparable to the existing image quality evaluation function. Although image fusion and image quality assessment based on deep learning and generative adversarial network have achieved good results, its \textit{objective function modeling is seriously restricted by the expression of this kind of cognition}. Although we combine the characteristics of human visual perception system in the image fusion task, there is still some gap with the complete human visual perception system, which is also a direction we need to study in the future.

\section{Conclusion}
Based on the characteristics of human visual perception system, we non-linear and selective of cross-modal image fusion method. \textit{\textbf{The most significant difference}} between our method and the current mainstream methods includes three points.
\textit{1)} We don't need a dedicated image fusion network to train first. 
\textit{2)} We introduce the illuminance fusion factor to simulate the non-linear characteristics of human visual perception for the first time in image fusion. 
\textit{3)} An attention mechanism was introduced in the image fusion task to simulate the selection characteristics of human visual perception. Through a large amount of data verification, experimental results demonstrate that our method is more in line with the human visual perception system than the existing mainstream method. 
 Although our algorithm does not fully simulate human visual perception characteristics, the first simulation of human visual perception characteristics in image fusion tasks is in line with the human visual perception mechanism. Although our method has achieved relatively good results compared with the existing algorithms, how to better learn the non-linear relationship between the features and the spatial structure will be further discussed in the next research work.

\begin{flushleft}	
\textbf{Acknowledgment}\\
\end{flushleft}
We are very grateful to Prof. Roundtree and Dr. Xiaoming Wang for their support of the language of the paper. This work was supported by the National Natural Science Foundation of China under Grants nos. 61871326, and the Shaanxi Natural Science Basic Research Program under Grant no. 2018JM6116.

\section*{References}

\bibliography{mybibfile}

\begin{thebibliography}{10}
\expandafter\ifx\csname url\endcsname\relax
  \def\url#1{\texttt{#1}}\fi
\expandafter\ifx\csname urlprefix\endcsname\relax\def\urlprefix{URL }\fi
\expandafter\ifx\csname href\endcsname\relax
  \def\href#1#2{#2} \def\path#1{#1}\fi

\bibitem{Treisman1980A}
A.~M. Treisman, G.~Gelade, A feature-integration theory of attention, Cognitive
  Psychology 12~(1) (1980) 97--136.

\bibitem{Valois1996Visual}
D.~Valois, K.~K., Visual perception: Foundations of vision, Science 271~(5254)
  (1996) 1371a--1371a.

\bibitem{Kubovy2003Foundations}
M.~Kubovy, W.~Epstein, S.~Gepshtein, Foundations of Visual Perception, John
  Wiley \& Sons, Inc., 2003.

\bibitem{Gallistel2009Memory}
N.~Johnstone, K.~Cohen~Kadosh, Why a developmental cognitive neuroscience
  approach may be key for future-proofing microbiota-gut-brain research,
  Behavioral and Brain Sciences 42 (2019) e73.
\newblock \href {http://dx.doi.org/10.1017/S0140525X18002753}
  {\path{doi:10.1017/S0140525X18002753}}.

\bibitem{Li2011Performance}
S.~Li, B.~Yang, J.~Hu, Performance comparison of different multi-resolution
  transforms for image fusion, Information Fusion 12~(2) (2011) 74--84.

\bibitem{Xiang2015A}
T.~Xiang, L.~Yan, R.~Gao, A fusion algorithm for infrared and visible images
  based on adaptive dual-channel unit-linking pcnn in nsct domain, Infrared
  Physics \& Technology 69 (2015) 53--61.

\bibitem{ZhangInfrared}
Z.~Xiaoye, M.~Yong, F.~Fan, Z.~Ying, H.~Jun, Infrared and visible image fusion
  via saliency analysis and local edge-preserving multi-scale decomposition,
  Journal of the Optical Society of America. A, Optics, Image Science, and
  Vision 34~(8) (2017) 1400--1410.

\bibitem{Ma2018Infrared}
J.~Ma, Y.~Ma, C.~Li, Infrared and visible image fusion methods and
  applications: A survey, Information Fusion 45 (2019) 153 -- 178.
\newblock \href
  {http://dx.doi.org/https://doi.org/10.1016/j.inffus.2018.02.004}
  {\path{doi:https://doi.org/10.1016/j.inffus.2018.02.004}}.

\bibitem{FLIR}
AZoSensors, Flir releases starter thermal imaging dataset for machine learning
  advanced driver assistance development (2018).

\bibitem{Zhang2013Dictionary}
Q.~Zhang, Y.~Fu, H.~Li, J.~Zou, Dictionary learning method for joint sparse
  representation-based image fusion, Optical Engineering 52~(5) (2013) 7006.

\bibitem{Ma2017InfraredWLS}
J.~Ma, Z.~Zhou, B.~Wang, H.~Zong, Infrared and visible image fusion based on
  visual saliency map and weighted least square optimization, Infrared Physics
  \& Technology 82 (2017) 8--17.

\bibitem{Liu2011LatentLATLRR}
G.~Liu, S.~Yan, Latent low-rank representation for subspace segmentation and
  feature extraction, in: International Conference on Computer Vision, 2011,
  pp. 1615--1622.
\newblock \href {http://dx.doi.org/10.1109/ICCV.2011.6126422}
  {\path{doi:10.1109/ICCV.2011.6126422}}.

\bibitem{Li2018InfraredLTLRR}
H.~Li, X.-J. Wu, J.~Kittler, Mdlatlrr: A novel decomposition method for
  infrared and visible image fusion, IEEE Transactions on Image Processing 29
  (2020) 4733--4746.

\bibitem{Li2018Infrared}
H.~Li, X.~jun Wu, T.~S. Durrani, Infrared and visible image fusion with resnet
  and zero-phase component analysis, Infrared Physics \$ Technology 102 (2019)
  103039.
\newblock \href
  {http://dx.doi.org/https://doi.org/10.1016/j.infrared.2019.103039}
  {\path{doi:https://doi.org/10.1016/j.infrared.2019.103039}}.

\bibitem{Liu2017InfraredCNN}
Y.~Liu, X.~Chen, J.~Cheng, H.~Peng, Z.~Wang, Infrared and visible image fusion
  with convolutional neural networks, International Journal of Wavelets
  Multiresolution \& Information Processing 16~(3) (2017) 1--20.

\bibitem{Nencini2007RemoteCVT}
F.~Nencini, A.~Garzelli, S.~Baronti, L.~Alparone, Remote sensing image fusion
  using the curvelet transform, Information Fusion 8~(2) (2007) 143--156.

\bibitem{Li_2018DL}
H.~Li, X.-J. Wu, J.~Kittler, Infrared and visible image fusion using a deep
  learning framework, 2018 24th International Conference on Pattern
  Recognition\href {http://dx.doi.org/10.1109/icpr.2018.8546006}
  {\path{doi:10.1109/icpr.2018.8546006}}.

\bibitem{Liu2015MultiDSIFT}
Y.~Liu, S.~Liu, Z.~Wang, Multi-focus image fusion with dense sift, Information
  Fusion 23~(C) (2015) 139--155.

\bibitem{MaFusionGAN}


\bibitem{Shutao2013ImageGF}
S.~Li, K.~Xudong, J.~Hu, Image fusion with guided filtering, IEEE Transactions
  on Image Processing 22~(7) (2013) 2864--2875.

\bibitem{Ma2016InfraredGTF}
J.~Ma, C.~Chen, C.~Li, J.~Huang, Infrared and visible image fusion via gradient
  transfer and total variation minimization, Information Fusion 31~(C) (2016)
  100--109.

\bibitem{Burt1987TheLP}
P.~J. Burt, E.~H. Adelson, The laplacian pyramid as a compact image code,
  Readings in Computer Vision 31~(4) (1987) 671--679.

\bibitem{Lahoud2019FastZERO}
F.~Lahoud, S.~Susstrunk, Zero-learning fast medical image fusion, in: 2019 22th
  International Conference on Information Fusion (FUSION), ISIF - International
  Society of Information Fusion, 2019, pp. 1--8.

\bibitem{Shreyamsha2015ImageCBF}
S.~Kumar, B.~K., Image fusion based on pixel significance using cross bilateral
  filter, Signal Image \& Video Processing 9~(5) (2015) 1193--1204.

\bibitem{Liu2016ImageCSR}
Y.~Liu, X.~Chen, R.~Ward, Z.~J. Wang, Image fusion with convolutional sparse
  representation, IEEE Signal Processing Letters 23~(12) (2016) 1882--1886.

\bibitem{Liu2017InfraredJSR-SD}
C.~Liu, Y.~Qi, W.~Ding, Infrared and visible image fusion method based on
  saliency detection in sparse domain, Infrared Physics \& Technology 83 (2017)
  94--102.

\bibitem{Liu2015ALPSR}
Y.~Liu, S.~Liu, Z.~Wang, A general framework for image fusion based on
  multi-scale transform and sparse representation, Information Fusion 24 (2015)
  147--164.

\bibitem{Naidu2011Image}
V.~P.~S. Naidu, Image fusion technique using multi-resolution singular value
  decomposition, Defence Science Journal 61~(5) (2011) 479--484.

\bibitem{Toet1989ImageRP}
A.~Toet, Image fusion by a ratio of low-pass pyramid, Pattern Recognition
  Letters 9~(4) (1989) 245--253.

\bibitem{Chipman1995Wavelets}
L.~J. Chipman, T.~M. Orr, L.~N. Graham, Wavelets and image fusion, in:
  International Conference on Image Processing, 1995.

\bibitem{Klonowski2008Importance}
W.~Klonowski, Importance of Nonlinear Signal Processing in Biomedicine,
  Springer Berlin Heidelberg, 2008.

\bibitem{Akay2000Nonlinear}
M.~Akay, Nonlinear Biomedical Signal Processing: Fuzzy Logic, Neural Networks,
  and New Algorithms, IEEE Press, 2000.

\bibitem{Zohary1992Population}
E.~Zohary, Population coding of visual stimuli by cortical neurons tuned to
  more than one dimension, Biological Cybernetics 66~(3) (1992) 265--272.

\bibitem{hu2017squeezeandexcitation}
J.~Hu, L.~Shen, G.~Sun, Squeeze-and-excitation networks, 2018.
\newblock \href {http://dx.doi.org/10.1109/CVPR.2018.00745}
  {\path{doi:10.1109/CVPR.2018.00745}}.

\bibitem{Zhang_2018RCAN}
Y.~Zhang, K.~Li, K.~Li, L.~Wang, B.~Zhong, Y.~Fu, Image super-resolution using
  very deep residual channel attention networks, Lecture Notes in Computer
  Science (2018) 294–310\href
  {http://dx.doi.org/10.1007/978-3-030-01234-2_18}
  {\path{doi:10.1007/978-3-030-01234-2_18}}.

\bibitem{Fu2018Dual}
J.~Fu, J.~Liu, H.~Tian, Y.~Li, Y.~Bao, Z.~Fang, H.~Lu, Dual attention network
  for scene segmentation (2018).
\newblock \href {http://arxiv.org/abs/1809.02983} {\path{arXiv:1809.02983}}.

\bibitem{Liu2017MultiCNN}
Y.~Liu, X.~Chen, H.~Peng, Z.~Wang, Multi-focus image fusion with a deep
  convolutional neural network, Information Fusion 36 (2017) 191--207.

\bibitem{Li2018DenseFuse}
H.~Li, X.~J. Wu, Densefuse: A fusion approach to infrared and visible images,
  IEEE Transactions on Image Processing 28~(5) (2018) 2614--2623.

\bibitem{fang2019crossmodal}
A.~Fang, X.~Zhao, Y.~Zhang, A cross-modal image fusion theory guided by human
  visual characteristics (2019).
\newblock \href {http://arxiv.org/abs/1912.08577} {\path{arXiv:1912.08577}}.

\bibitem{MA202085}
J.~Ma, P.~Liang, W.~Yu, C.~Chen, J.~Jiang, Infrared and visible image fusion
  via detail preserving adversarial learning, Information Fusion 54 (2019)
  85--98.

\bibitem{9031751}
J.~Ma, H.~Xu, J.~Jiang, X.~Mei, X.-P. Zhang, Ddcgan: A dual-discriminator
  conditional generative adversarial network for multi-resolution image fusion
  29 (2020) 1--1.

\bibitem{Bavirisetti2016Two}
D.~P. Bavirisetti, R.~Dhuli, Two-scale image fusion of visible and infrared
  images using saliency detection, Infrared Physics \& Technology 76 (2016)
  52--64.

\bibitem{Zhou2016Fusion}
J.~Zhu, W.~Jin, L.~Li, Z.~Han, X.~Wang, Fusion of the low-light-level visible
  and infrared images for night-vision context enhancement, Chinese Optics
  Letters 16~(1) (2018) 650--667.

\bibitem{li2018infraredlrr}
H.~Li, X.-J. Wu, Infrared and visible image fusion using latent low-rank
  representation (2018).
\newblock \href {http://arxiv.org/abs/1804.08992} {\path{arXiv:1804.08992}}.

\bibitem{TNO}
A.~Toet, Tno dataset (2018).

\bibitem{Summers2003Harvard}
Summers, D, Harvard whole brain atlas, Journal of Neurology Neurosurgery \&
  Psychiatry 74~(3) (2003) 288--288.

\bibitem{zhang2020vifb}


\bibitem{1576816}
H.~R. Sheikh, A.~C. Bovik, Image information and visual quality, IEEE
  Transactions on Image Processing 15~(2) (2006) 430--444.
\newblock \href {http://dx.doi.org/10.1109/TIP.2005.859378}
  {\path{doi:10.1109/TIP.2005.859378}}.

\bibitem{Cui2015Detail}
G.~Cui, H.~Feng, Z.~Xu, Q.~Li, Y.~Chen, Detail preserved fusion of visible and
  infrared images using regional saliency extraction and multi-scale image
  decomposition, Optics Communications 341~(341) (2015) 199--209.

\bibitem{1284395}
Z.~Wang, A.~C. Bovik, H.~R. Sheikh, E.~P. Simoncelli, Image quality assessment:
  from error visibility to structural similarity, IEEE Transactions on Image
  Processing 13~(4) (2004) 600--612.
\newblock \href {http://dx.doi.org/10.1109/TIP.2003.819861}
  {\path{doi:10.1109/TIP.2003.819861}}.

\bibitem{Qu2002Information}
G.~Qu, D.~Zhang, P.~Yan, Information measure for performance of image fusion,
  Electronics Letters 38~(7) (2002) 313--315.

\bibitem{Han2013A}
Y.~Han, Y.~Cai, Y.~Cao, X.~Xu, A new image fusion performance metric based on
  visual information fidelity, Information Fusion 14~(2) (2013) 127--135.

\bibitem{Sheikh2006An}
H.~R. Sheikh, Member, IEEE, A.~C. Bovik, Fellow, An information fidelity
  criterion for image quality assessment using natural scene statistics, IEEE
  Trans Image Process 14~(12) (2006) 2117--2128.

\bibitem{ZHANG202099}
Y.~Zhang, Y.~Liu, P.~Sun, H.~Yan, X.~Zhao, L.~Zhang, Ifcnn: A general image
  fusion framework based on convolutional neural network, Information Fusion 54
  (2020) 99 -- 118.
\newblock \href
  {http://dx.doi.org/https://doi.org/10.1016/j.inffus.2019.07.011}
  {\path{doi:https://doi.org/10.1016/j.inffus.2019.07.011}}.

\bibitem{Lewis2007Pixel}
J.~J. Lewis, R.~J. O’Callaghan, S.~G. Nikolov, D.~R. Bull, N.~Canagarajah,
  Pixel- and region-based image fusion with complex wavelets, Information
  Fusion 8~(2) (2007) 119--130.

\bibitem{Durga2019Multi}
D.~Bavirisetti, G.~Xiao, J.~Zhao, R.~Dhuli, G.~Liu, Multi-scale guided image
  and video fusion: A fast and efficient approach, Circuits, Systems, and
  Signal Processing 38~(12) (2019) 5576--5605.

\bibitem{Bavirisetti2016Fusion}
D.~P. Bavirisetti, R.~Dhuli, Fusion of infrared and visible sensor images based
  on anisotropic diffusion and karhunen-loeve transform, IEEE Sensors Journal
  16~(1) (2016) 203--209.

\bibitem{Bavirisetti2017Multi}
D.~P. Bavirisetti, Multi-sensor image fusion based on fourth order partial
  differential equations, in: 20th International Conference on Information
  Fusion, 2017.

\bibitem{Zhang2017Infrared}
Y.~Zhang, L.~Zhang, X.~Bai, L.~Zhang, Infrared and visual image fusion through
  infrared feature extraction and visual information preservation, Infrared
  Physics and Technology 83 (2017) 227--237.

\bibitem{Liu2011Latent}
G.~Liu, S.~Yan, Latent low-rank representation for subspace segmentation and
  feature extraction, in: International Conference on Computer Vision, 2011.

\bibitem{ZhaoT.2019Pfan}
T.~Zhao, X.~Wu, Pyramid feature attention network for saliency detection, Vol.
  2019, IEEE Computer Society, 2019, pp. 3080--3089.

\bibitem{Zhou2015Perceptual}
Z.~Zhou, B.~Wang, S.~Li, M.~Dong, Perceptual fusion of infrared and visible
  images through a hybrid multi-scale decomposition with gaussian and bilateral
  filters, Information Fusion 30 (2016) 15--26.

\end{thebibliography}

\end{document}